\documentclass[journal]{IEEEtran}

\ifCLASSINFOpdf
   \usepackage[pdftex]{graphicx}
\else
   \usepackage[dvips]{graphicx}
\fi

\usepackage{amsmath}
\usepackage{amssymb}
\usepackage{mathrsfs}
\usepackage{algorithm}
\usepackage{algorithmic}
\usepackage{subfigure}
\usepackage{xcolor}
\usepackage{bm}

\begin{document}
\title{DR$^{2}$-Net: \textbf{D}eep \textbf{R}esidual \textbf{R}econstruction \textbf{N}etwork for Image Compressive Sensing}
\author{
Hantao Yao,
Feng Dai,
Dongming Zhang,
Yike Ma,
Shiliang Zhang, \emph{Member, IEEE},
Yongdong Zhang, \emph{Member, IEEE},
Qi Tian, \emph{Fellow, IEEE}

\IEEEcompsocitemizethanks{
\IEEEcompsocthanksitem Hantao Yao is with Key Lab of Intelligent Information Processing of Chinese Academy of Sciences (CAS), Institute of Computing Technology, CAS, Beijing 100190, China, and also with the University of the Chinese Academy of Sciences, Beijing 100049, China, Email: yaohantao@ict.ac.cn
\IEEEcompsocthanksitem Feng Dai and Yike Ma is with Key Lab of Intelligent Information Processing of Chinese Academy of Sciences (CAS), Institute of Computing Technology, CAS, Beijing 100190, China,
\IEEEcompsocthanksitem Dongming Zhang National Computer Network Emergency Response Technical Team/Coordination Center of China
\IEEEcompsocthanksitem Shiliang Zhang is with School of Electronic Engineering and Computer Science, Peking University, Beijing 100871, China, Email: slzhang.jdl@pku.edu.cn
\IEEEcompsocthanksitem Yongdong Zhang, is with Key Lab of Intelligent Information Processing of Chinese Academy of Sciences (CAS), Institute of Computing Technology, CAS, Beijing 100190, China, and also with the Beijing Advanced Innovation Center for Imaging Technology, Capital Normal University, Beijing 100048, China, Email: zhyd@ict.ac.cn
\IEEEcompsocthanksitem Qi Tian is with Department of Computer Science University of Texas at San Antonio, San Antonio, USA, Email: qitian@cs.utsa.edu
}}

\maketitle

\begin{abstract}
Most traditional algorithms for compressive sensing image reconstruction suffer from the intensive computation. Recently, deep learning-based reconstruction algorithms have been reported, which dramatically reduce the time complexity than iterative reconstruction algorithms. In this paper, we propose a novel \textbf{D}eep \textbf{R}esidual \textbf{R}econstruction Network (DR$^{2}$-Net) to reconstruct the image from its Compressively Sensed (CS) measurement. The DR$^{2}$-Net is proposed based on two observations: 1) linear mapping could reconstruct a high-quality preliminary image, and 2) residual learning could further improve the reconstruction quality. Accordingly, DR$^{2}$-Net consists of two components, \emph{i.e.,} linear mapping network and residual network, respectively. Specifically, the fully-connected layer in neural network implements the linear mapping network. We then expand the linear mapping network to DR$^{2}$-Net by adding several residual learning blocks to enhance the preliminary image. Extensive experiments demonstrate that the DR$^{2}$-Net outperforms traditional iterative methods and recent deep learning-based methods by large margins at measurement rates 0.01, 0.04, 0.1, and 0.25, respectively. The code of DR$^{2}$-Net has been released on: https://github.com/coldrainyht/caffe\_dr2
\end{abstract}


\IEEEpeerreviewmaketitle

\section{Introduction}\label{Sect:Intro}
\begin{figure}
\begin{center}
\subfigure[]{
\includegraphics[width=1\linewidth]{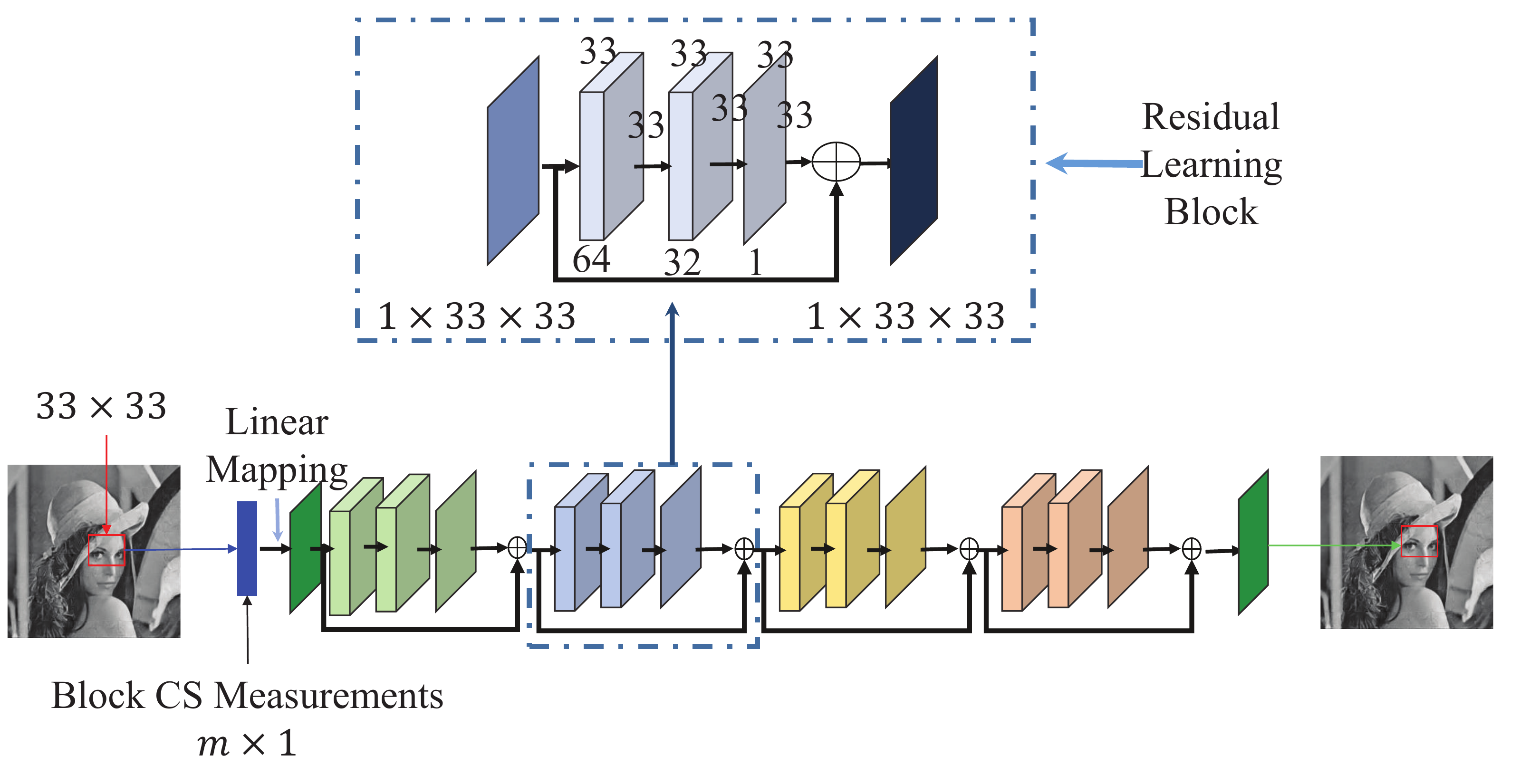}}
\subfigure[]{
\includegraphics[width=1\linewidth]{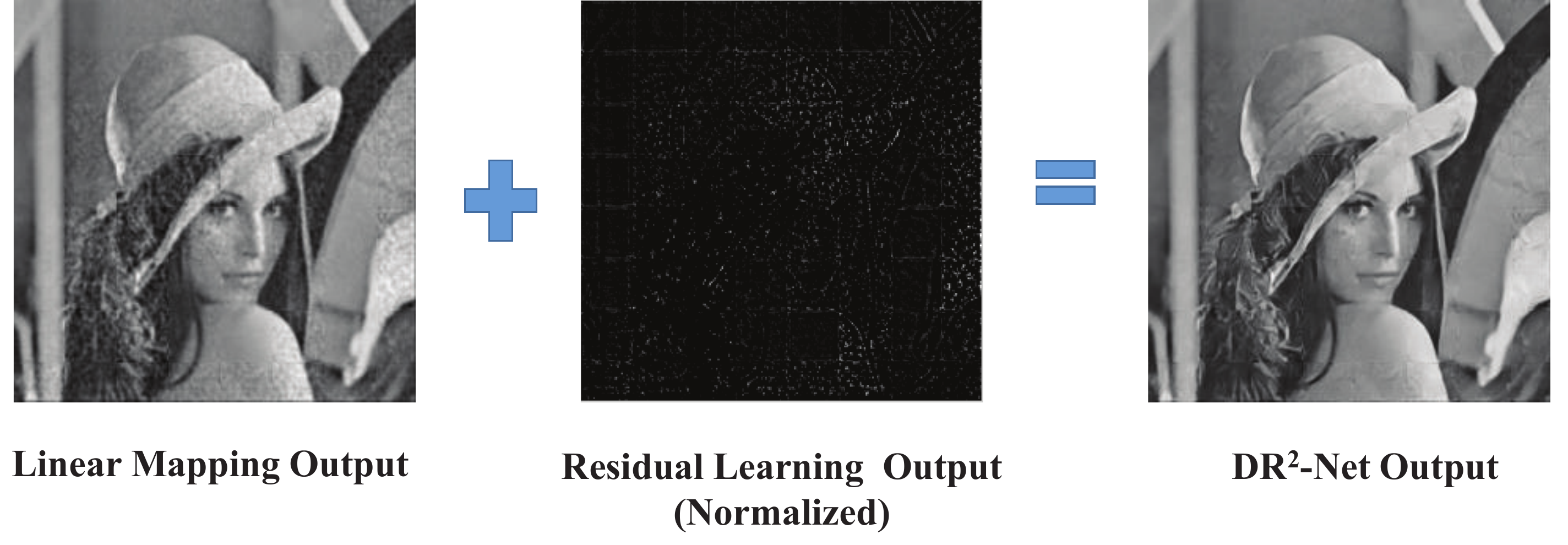}}
\end{center}
\caption{(a) The framework of Deep Residual Reconstruction Network (DR$^{2}$-Net). (b) Illustration of the two components for image reconstruction in DR$^{2}$-Net. The right most image is the output of DR$^{2}$-Net (Best viewed in color pdf).}
\label{Fig:framework}
\end{figure}
Compressive sensing is an emerging technique to acquire and reconstruce digital data, \emph{e.g.,} images and videos. It is to capture the data in the form of Compressively Sensed (CS) measurements, then reconstruct the original data from these CS measurements. Because the required number of measurements is far fewer than limited by the Nyquist theory, compressive sensing is very desirable in many applications such as cameras, medical scanners and so on.

By achieving perfect image reconstruction from CS measurements, some algorithms have been proposed in recently years~\cite{daubechies2004iterative,li2013efficient,mairal2009non,mallat1993matching,tropp2007signal}. Most of these approaches commonly model the image signal with structured sparsity assumption and solve an optimization problem with iterative optimization strategies. The intensive computation of this iterative optimization has become the bottleneck for the application of image compressive sensing.

Deep neural networks~\cite{krizhevsky2012imagenet} have exhibited a series of breakthroughs in computer vision tasks, \emph{e.g.,} image classification~\cite{krizhevsky2012imagenet}, super-resolution~\cite{dong2014learning}, and restoration~\cite{svoboda2016compression}, \emph{etc.} Recently, several deep neural networks have been proposed for compressive sensing image reconstruction. Thanks to the powerful learning ability, current deep learning-based methods effectively avoid the expensive computation in traditional approaches and achieve a promising reconstruction performance~\cite{kulkarni2016reconnet,mousavi2015deep}.

This paper proposes a Deep Residual Reconstruction Network (DR$^{2}$-Net), which further boosts the reconstruction quality from Compressively Sensed (CS) measurements with fast speed. The DR$^{2}$-Net takes the CS measurement of an image patch as input and consists of two components: linear mapping network and residual network. The linear mapping network is first used to obtain a preliminary reconstructed image, the residual network is then applied to infer the residual between the ground truth image and the preliminary reconstructed image. The residual finally updates the preliminary result as the output of DR$^{2}$-Net. An intuitive description of DR$^{2}$-Net is illustrated in Fig.~\ref{Fig:framework}.

Extensive experiments show our DR$^{2}$-Net significantly outperforms existing iterative methods and deep learning-based methods. Our contributions could be summarized as follows:
\begin{itemize}
  \item We propose a novel Deep Residual Reconstruction Network (DR$^{2}$-Net) for image compressive sensing, which outperforms the existing works. With DR$^{2}$, the deep-learning CS methods outperform traditional methods at all four measurement rates for the first time.
  \item The linear mapping generates a reasonably good preliminary reconstruction image with faster speed and lower computational cost. The linear mapping network is well-suited to image reconstruction task in computational resource and network bandwidth limited scenarios.
  \item The residual learning further improves the reconstruction quality. The residual learning is independent with the linear mapping stage. Therefore, the existing deep learning-based methods could flexibly fuse with residual learning to chase higher image reconstruction quality.
\end{itemize}

\section{Related Work}\label{Sect:RW}

This work is related with image compressive sensing image reconstruction, deep learning for compressive sensing, and the deep residual network. In the following, we will review these three categories of related works.

\emph{Compressive sensing image reconstruction: } Compressive sensing is an emerging technique to acquire and process digital data like images and videos. It aims to recover the source signal $\mathbf{x}\in \mathbf{R}^{n\times 1}$ from the randomized CS measurements, \emph{i.e.,} $\mathbf{y}=\mathbf{\Phi} \mathbf{x} (m\ll n)$, where $\Phi\in \mathbf{R}^{m\times n}$, and $\mathbf{y}\in \mathbf{R}^{m\times 1}$. Since $m\ll n$, the equation has multiple solutions. Thus, there exists more than one $\mathbf{x}\in \mathbf{R}^{n\times 1}$ that can yield the same CS measurements $\mathbf{y}$. To reconstruct the original signals from CS measurements, the early recovery algorithms assume the original image signal has $l_{p}$-norm ($0\le p\le 1$) sparsity. Based on this assumption, researchers propose several iterative algorithms, \emph{e.g.,} matching pursuit~\cite{mallat1993matching}, orthogonal matching pursuit~\cite{tropp2007signal}, iterative hard-thresholding~\cite{blumensath2009iterative}, compressive sampling matching search~\cite{needell2010cosamp}, approximate message passing~\cite{donoho2009message}, and iterative soft-thresholding~\cite{daubechies2004iterative}, \emph{etc}. Furthermore, many elaborate structures, such as total variation sparsity~\cite{xiao2012alternating,li1tval3}, non-local sparsity~\cite{mairal2009non,dong2012image,mairal2009non}, block sparsity~\cite{eldar2010block}, wavelet tree sparsity~\cite{baraniuk2010model,hegde2014fast} are used as prior knowledge of the original image signals, which are beneficial to compressive sensing image reconstruction. All these recovery algorithms solve an optimization problem. The iterative solvers are often used and lead to expensive computation, which has become the bottleneck for the application of image compressive sensing. Recently, some deep learning-based approaches have been proposed for this problem. Stacked Denoising Auto-encoders (SDAs)~\cite{mousavi2015deep} and Convolutional Neural Network (CNN)~\cite{kulkarni2016reconnet} are applied to compressive sensing image reconstruction. Existing works~\cite{mousavi2015deep,kulkarni2016reconnet} have shown that the deep learning-based approaches are about 3 orders of magnitude faster than traditional reconstruction algorithms. 

\emph{Deep learning for compressive sensing:} Deep learning has exhibited promising performance in computer vision and image processing tasks, \emph{e.g.,} semantic segmentation~\cite{long2015fully}, depth estimation~\cite{eigen2014depth}, image super-resolution~\cite{dong2014learning} and image denoising~\cite{xie2012image}. Recently, some works introduce the deep learning into image/video reconstruction. For compressive image recovery, Kulkarni~\emph{etal.}~\cite{kulkarni2016reconnet} propose a CNN-based ReconNet to achieve non-iterative compressive sensing image reconstruction. In~\cite{mousavi2015deep}, Stacked Denoising Auto-encoders (SDAs) are employed to learn a mapping between CS measurements and image blocks. For compressive video recovery, Iliadis~\emph{etal.}~\cite{iliadis2016deepbinarymask, iliadis2016deep} propose Deep Fully-Connected Network, where the encoder learns binary sensing mask and the decoder determines the reconstruction of the video. These approaches effectively avoid the expensive computation in traditional approaches and have achieved promising image/video reconstruction performance.

\emph{Deep Residual Network:} Recently, the deep Residual Network (ResNets)~\cite{he2015deep} has achieved promising performance on several computer vision tasks. Compared with the traditional Convolutional Neural Network, \emph{e.g.,} AlexNet~\cite{krizhevsky2012imagenet}, VGG~\cite{simonyan2014very}, and GoogleNet~\cite{szegedy2015going}, the ResNets introduces identity shortcut connections that directly pass the data flow to later layers, thus effectively avoids signal attenuation caused by multiple stacked non-linear transformations. As a consequence, deeper network can be constructed with ResNets and faster training speed can be achieved. By going deeper in the network, ResNet generally gets better performance in comparison with other deep learning models.

Among the related works, the most similar work to ours is the one by Kulkarni~\emph{etal.}~\cite{kulkarni2016reconnet}, who propose a ReconNet for compressive sensing image reconstruction. However, the proposed Deep Residual Reconstruction Network (DR$^{2}$-Net) is different from the one in~\cite{kulkarni2016reconnet} in the aspects of both motivation and network structure. The method in~\cite{kulkarni2016reconnet} is the first work that employ CNN for compress sensing. The design of ReconNet~\cite{kulkarni2016reconnet} is inspired by the SRCNN~\cite{}, and does not take the essential of CS into consideration. Therefore, it is not a carefully-designed network for CS image reconstruction. By give more analysis for compressive sensing problem, we found that simply using the linear mapping (fully-connected operation) could perfectly suitable to solve the CS, and it is not a optimal choice to train ReconNet in an end-to-end way. The DR$^{2}$-Net firstly trains a linear mapping network to obtain a reasonably good preliminary reconstruction image, then trains the residual network to further improve the reconstruction quality. We find that this strategy is more suitable for applying CNN on image compressive sensing.

\section{Deep Residual Reconstruction Network} \label{Sect:DRR}

As shown in Fig.~\ref{Fig:framework}(a), DR$^{2}$-Net takes the Compressively Sensed (CS) measurements of the $33\times 33$ sized image patch as input, and outputs a $33\times 33$ reconstructed image. The DR$^{2}$-Net contains a linear mapping sub-network $\mathcal{F}^{f}(\cdot)$ and a residual sub-network $\mathcal{F}^{r}(\cdot)$. $\mathcal{F}^{f}(\cdot)$ generates a preliminary reconstructed image. $\mathcal{F}^{r}(\cdot)$ infers the residual between the preliminary image and ground truth image. In the following, we firstly give details of linear mapping and residual learning. Then, we introduce the architecture of DR$^{2}$-Net. Finally, we summarize the procedure that DR$^{2}$-Net reconstructs an image.

\subsection{Linear Mapping}\label{Sect:LM}

Compressive sensing signal reconstruction recovers the signal $\mathbf{x}\in\mathbf{R}^{n\times 1}$ from its CS measurement $\mathbf{y}\in\mathbf{R}^{m\times 1}$. The mapping from $\mathbf{y}$ to $\mathbf{x}$ can be regarded as an approximate linear mapping, \emph{i.e.},
\begin{equation}
\mathbf{x}=\mathbf{W}\mathbf{y},
\label{Eq:mapping}
\end{equation}
where $\mathbf{W}\in \mathbf{R}^{n\times m}$ is the mapping matrix. Eq.~\eqref{Eq:mapping} is an overdetermined equation, therefore it has no accurate solution. However, we could estimate a mapping matrix $\mathbf{W}^{f}$, which makes $||\mathbf{x}-\mathbf{W}^{f}\mathbf{y}||^{2}_{2}$ have the minimum error.

We denote the CS measurement and its corresponding source signal as $\mathbf{y}_{i}\in \mathbf{R}^{m\times 1}$ and $\mathbf{x}_{i}\in \mathbf{R}^{n\times 1}$, respectively. The training data containing $N$ training samples can be denoted as $\{(\mathbf{y}_{1},\mathbf{x}_{1}),  (\mathbf{y}_{2},\mathbf{x}_{2}),......,(\mathbf{y}_{N},\mathbf{x}_{N})\}$. Based on this training data, we could get $\mathbf{W}^{f}$ by solving the following optimization, \emph{i.e.},
\begin{equation}
	\mathbf{W}^{f}=arg\min_{\mathbf{W}} ||\mathbf{X}-\mathbf{W}\mathbf{Y}||_{2}^{2},
\label{Eq:2}
\end{equation}
where $\mathbf{X}=[\mathbf{x}_{1},\mathbf{x}_{2},......,\mathbf{x}_{N}]$, and $\mathbf{Y}=[\mathbf{y}_{1},\mathbf{y}_{2},......,\mathbf{y}_{N}]$.

Eq.~\eqref{Eq:2} presents a linear mapping function, which could be effectively simulated by fully-connected layer in deep learning. Consequently, we employ the linear mapping network $\mathcal{F}^{f}(\cdot)$ composed of fully connected layer to infer the optimal mapping matrix $\mathbf{W}^{f}$ with the loss function in Eq.~\eqref{Eq:4}, \emph{i.e.},
\begin{equation}
L(\{\mathbf{W}^{f}\})=\frac{1}{N}\sum_{i=1}^{N}||\mathbf{x}_{i}-\mathcal{F}^{f}(\mathbf{y}_{i},\mathbf{W}^{f})||_{2}^{2},
\label{Eq:4}
\end{equation}
where $\mathcal{F}^{f}(\mathbf{y}_{i},\mathbf{W}^{f})$ is the output of linear mapping network for input $\mathbf{y}_{i}$.

The network $\mathcal{F}^{f}(\cdot)$ contains one fully-connected layer with 1,089 neurons. By using Stochastic Gradient Descent (SGD) training algorithm, we could obtain an optimal mapping matrix $\mathbf{W}^{f}$ corresponding to the minimum Eq.~\eqref{Eq:2} on all training samples.

With the trained linear mapping network $\mathcal{F}^{f}(\cdot)$, a preliminary reconstructed image $\widehat{\mathbf{x}_{i}}$ could be obtained for any given CS measurement $\mathbf{y}_{i}$. We denote the procedure of generating the preliminary reconstructed image as
\begin{equation}
\widehat{\mathbf{x}_{i}}=\mathcal{F}^{f}(\mathbf{y}_{i},\mathbf{W}^{f}).
\label{Eq:recon}
\end{equation}

\subsection{Residual Learning}\label{Sect:RL}

The linear mapping network is trained to chase an approximate solution to the source signal $\mathbf{x}_{i}$, because it is hard to get an accurate solution for Eq.~\eqref{Eq:mapping}. To further narrow down the gap between $\widehat{\mathbf{x}_{i}}$ and $\mathbf{x}_{i}$, the residual learning network is introduced to estimate the gap between the two signals. In other words, the next step is to infer the residual $\mathbf{d}_{i}$ based on $\widehat{\mathbf{x}_{i}}$, \emph{i.e.},
\begin{equation}
\mathbf{d}_{i} = \mathbf{x}_{i}-\widehat{\mathbf{x}_{i}}.
\label{Eq:diff}
\end{equation}

Residual learning block is introduced in this part to estimate the residual $\mathbf{d}_{i}$ from $\widehat{\mathbf{x}_{i}}$. As illustrated in Fig.~\ref{Fig:framework}(b), the output of residual learning is fused with the output of linear mapping as the final image construction result.

We firstly give a brief introduction to residual learning. Given an input $\mathbf{\mathcal{X}}$, we denote its desired underlying mapping performed by a few stacked layers in neural network as $\mathcal{H}(\mathbf{\mathcal{X}})$. Most traditional methods expect to directly learn the mapping $\mathcal{H}(\cdot)$, \emph{e.g.,} using several convolutional operations to fit $\mathcal{H}(\cdot)$~\cite{simonyan2014very}. If the output $\mathcal{H}(\mathbf{\mathcal{X}})$ is similar to $\mathbf{\mathcal{X}}$, we expect the $\mathcal{H}(\cdot)$ is an identity mapping. However, it is difficult to optimize $\mathcal{H}(\cdot)$ as the identity mapping in practice. On the other hand, for the case that the $\mathcal{H}(\mathbf{\mathcal{X}})$ is similar to $\mathbf{\mathcal{X}}$, the residuals between $\mathcal{H}(\mathbf{\mathcal{X}})$ and $\mathbf{\mathcal{X}}$ would be closed to zero. As the weights of convolutional layers are always initialized to have zero-means, the convolutional layers could be easily trained to approximate the residual $\mathcal{F}(\mathbf{\mathcal{X}})$ in Eq.~\eqref{Eq:res}. Therefore, rather than expect stacked layers to approximate $\mathcal{H}(\cdot)$, we explicitly let these layers approximate the residual function in Eq.~\eqref{Eq:res}. More details of residual learning can be found in~\cite{he2015deep}.

\begin{equation}
 \mathcal{F}(\mathbf{\mathcal{X}}):= \mathcal{H}(\mathbf{\mathcal{X}})-\mathbf{\mathcal{X}}.
 \label{Eq:res}
\end{equation}

As a consequence, the convolutional layers initialised with zero-means could be trained to estimate the residual $\mathbf{d}_{i}$ in Eq.~\eqref{Eq:diff}. We implement the residual learning with residual learning blocks shown in Fig.~\ref{Fig:framework}(a). Each block contains three convolution layers to simulate the complicated non-linear mappings. The residual network $\mathcal{F}^{r}(\cdot)$ consists of several residual learning blocks.

Specifically, with $\widehat{\mathbf{x}_{i}}$ as input, residual network $\mathcal{F}^{r}( \widehat{\mathbf{x}_{i}},\mathbf{W}^{r})$ generates an estimated residual $\widehat{\mathbf{d}_{i}}$, where $\mathbf{W}^{r}$ is the parameters for residual network. We denote this procedure as
\begin{equation}
\widehat{\mathbf{d}_{i}}=\mathcal{F}^{r}( \widehat{\mathbf{x}_{i}},\mathbf{W}^{r}).
\label{Eq:residual_network}
\end{equation}

The DR$^{2}$-Net takes the CS measurement $\mathbf{y}_{i}$ as input, first obtains the preliminary reconstructed image $\widehat{\mathbf{x}_{i}}$, then fuses it with the estimated residual $\widehat{\mathbf{d}_{i}}$ from residual network as the final result, \emph{i.e.},
\begin{equation}
\mathbf{x}^{*}_{i}=\widehat{\mathbf{x}_{i}} + \widehat{\mathbf{d}_{i}}.
\end{equation}

By replacing the $\widehat{\mathbf{x}_{i}}$ and $\widehat{\mathbf{d}_{i}}$ with Eq.~\eqref{Eq:recon} and Eq.~\eqref{Eq:residual_network}, the final $\mathbf{x}^{*}_{i}$ is obtained by two networks, \emph{i.e.},
\begin{equation}
\mathbf{x}^{*}_{i}=\mathcal{F}^{f}(\mathbf{y}_{i},\mathbf{W}^{f}) + \mathcal{F}^{r}(\mathcal{F}^{f}(\mathbf{y}_{i},\mathbf{W}^{f}) ,\mathbf{W}^{r}).
\label{Eq:drloss}
\end{equation}

The DR$^{2}$-Net is trained with SGD according to the loss function in Eq.~\eqref{Eq:loss}. The training procedure
obtains the optimal parameters $\mathbf{W}^{f}$ and $\mathbf{W}^{r}$, respectively.
\begin{equation}
L(\{\mathbf{W}^{f},\mathbf{W}^{r}\})=\frac{1}{N}\sum_{i=1}^{N}||\mathbf{x}_{i}-\mathbf{x}^{*}_{i}||_{2}^{2}.
\label{Eq:loss}
\end{equation}

\subsection{DR$^{2}$-Net Architecture}\label{Sect:dr2_architecture}

DR$^{2}$-Net processes the CS measurement computed from a $33\times 33$ sized image block. The linear mapping network $\mathcal{F}^{f}(\cdot)$ takes the CS measurement $\mathbf{y}\in \mathbf{R}^{m\times 1}$ as input, where $m$ is various with Measurement Rates (MRs), \emph{e.g.,} $m=$272, 109, 43, and 10 corresponding to MR = 0.25, 0.10, 0.04, and 0.01, respectively. The linear mapping network consists of one fully-connected layer with $1089$ neurons. We reshape its $1089$-dim output to the size of $33\times 33$ as the preliminary reconstructed image.

As shown in Fig.~\ref{Fig:framework}(a), the residual network $\mathcal{F}^{r}(\cdot)$ takes preliminary result as input, and outputs one image with size $33\times 33$. The residual network contains four residual learning blocks and each block contains three convolutional layers. In each block, the first convolution layer uses kernel of size $11\times 11$ and generates $64$ feature maps. The second layer produces $32$ features maps with $1\times 1$ kernel. The third layer generates $1$ feature map with $7\times 7$ kernel. To generate the feature map with size $33\times 33$, we add corresponding padding on each layer. Inspired by previous work~\cite{he2015deep}, we further add batch normalization~\cite{ioffe2015batch} to the first two convolutional layers of each block. Each convolutional layer is followed by a ReLU layer except the last convolutional layer.

\subsection{Reconstruction Procedure}

Given an image, we firstly divide it into $33\times 33$ image patches with no overlap, and extract the CS measurement on each patch. Next, DR$^{2}$-Net takes the CS measurements as input and outputs the reconstructed patches. The reconstructed patches compose an intermediate reconstructed image, which is finally processed with BM3D~\cite{dabov2007image} to remove the artifacts caused by block-wise processing.

\section{DR$^{2}$-Net Training } \label{sec:NT_Train}
\subsection{Training Data Generation}

We use the same set of 91 images used in~\cite{kulkarni2016reconnet} to generate the training data for  DR$^{2}$-Net. We first resize the original images to three scales \emph{i.e.,} 0.75, 1, and 1.5, to construct a scale space. Then, on each scale, we extract $33\times 33$ sized image patches with stride 14. This process finally samples 86,656 image patches from 273 images. For each image patch, we firstly extract its luminance component, denoting the luminance component as $\mathbf{x}_{i}$ and then compute its CS measurement $\mathbf{y}_{i}=\mathbf{\Phi} \mathbf{x}_{i}$, where $\mathbf{\Phi}$ is a random Gaussian matrix. $(\mathbf{y}_{i}, \mathbf{x}_{i})$ is thus an input-output pair for DR$^{2}$-Net training, where $\mathbf{y}_{i}$ is the network input and $\mathbf{x}_{i}$ is its ground truth label.

\subsection{Training Strategy}

The training procedure of DR$^{2}$-Net consists of two steps for two sub-networks $\mathcal{F}^{f}(\cdot)$ and $\mathcal{F}^{r}(\cdot)$, respectively. The first step uses a relatively large learning rate with step strategy to train the network $\mathcal{F}^{f}(\cdot)$. Parameters like maximum number of iterations, learning rate, stepsize, and gamma are set as 1,000,000, 0.001, 200,000, and 0.5, respectively. With the trained $\mathbf{W}^{f}$, the second step trains DR$^{2}$-Net in an end-to-end manner in up to 100,000 iterations with a smaller fixed learning rate set as 0.00001. Note that, the second step updates both the $\mathbf{W}^{f}$ and $\mathbf{W}^{r}$.

\begin{table*}
\footnotesize
\caption{PSNR values in dB for testing images  by different algorithms at different Measurement Rates. For TVAL3, NLR-CS, D-AMP, and SDA, we use the results reported in~\cite{kulkarni2016reconnet}. The reconstruction results for those  images are shown in Fig.~\ref{Fig:reconstruction_results}. ``w/o BM3D'' means without applying BM3D, and ``w/ BM3D'' denotes using BM3D. ``\textbf{Mean PSNR}" is the mean PSNR value among all 11 testing images.}
\centering
\begin{tabular}{|l|c|c|c|c|c|c|c|c|c|}
\hline
Image Name & Methods &\multicolumn{2}{|c|}{MR=0.25}&\multicolumn{2}{|c|}{MR=0.10}&\multicolumn{2}{|c|}{MR=0.04}&\multicolumn{2}{|c|}{MR=0.01}\\
\cline{3-10}
& & w/o BM3D & w/ BM3D& w/o BM3D & w/ BM3D& w/o BM3D & w/ BM3D& w/o BM3D & w/ BM3D \\
\hline
	& TVAL3~\cite{li2013efficient} & 24.19&24.20 &21.88 &22.21& 18.98& 18.98&11.94&11.96  \\
	& NLR-CS~\cite{dong2014compressive} & \textbf{28.01}&\textbf{28.00}& 14.80& 14.84&11.08 &11.56 &5.50&5.86  \\
Barbara	& D-AMP~\cite{metzler2014denoising} &25.08 &25.96 &21.23 & 21.23&16.37 & 16.37&5.48&  5.48\\
	& SDA~\cite{mousavi2015deep} &23.19 & 23.20& 22.07& 22.39& 20.49& 20.86&18.59& 18.76 \\
	& ReconNet~\cite{kulkarni2016reconnet} & 23.25& 23.52& 21.89& 22.50&20.38 & 21.02&18.61& 19.08 \\
	& DR$^{2}$-Net &25.77 & 25.99&\textbf{22.69}&\textbf{22.82} &\textbf{20.70}& \textbf{21.30}&\textbf{18.65}&\textbf{19.10}  \\	
\hline
	& TVAL3 & 22.70&22.71 &18.69&18.70& 16.04& 16.05&10.35&10.37  \\
	& NLR-CS & 23.52&23.52 &12.81 & 12.83& 9.66&10.10 &4.85& 5.18 \\
Fingerprint	& D-AMP &25.17 & 23.87&17.15 &16.88 & 13.82& 14.00&4.66& 4.73 \\
	& SDA & 24.28&23.45 & 20.29&20.31 & 16.87& 16.83&\textbf{14.83}&  14.82\\
	& ReconNet &25.57 & 25.13& 20.75& 20.97&16.91 &16.96 &14.82& 14.88 \\
	& DR$^{2}$-Net & \textbf{27.65}& \textbf{27.75}&\textbf{22.03}&\textbf{22.45} &\textbf{17.40}&\textbf{17.47} &14.73&\textbf{14.95}  \\	
\hline
	& TVAL3 & 24.05&24.07 &18.88 &18.92& 14.88& 14.91&9.75&9.77  \\
	& NLR-CS & 22.43&22.56&12.18&12.21&8.96&9.29&4.45&4.77 \\
Flinstones	& D-AMP &25.02&24.45&16.94&16.82&12.93&13.09&4.33&4.34\\
	& SDA &20.88&20.21&18.40&18.21&16.19&16.18& 13.90&13.95\\
	& ReconNet &22.45&22.59&18.92&19.18&16.30&16.56&13.96&14.08 \\
	& DR$^{2}$-Net &\textbf{26.19}&\textbf{26.77} &\textbf{21.09}&\textbf{21.46}&\textbf{16.93}&\textbf{17.05} &\textbf{14.01}&\textbf{14.18} \\	
\hline
	& TVAL3 &28.67&28.71&24.16&24.18&19.46&19.47&11.87&11.89 \\
	& NLR-CS &29.39&\textbf{29.67}&15.30&15.33&11.61&11.99&5.95&6.27 \\
Lena	& D-AMP &28.00&27.41&22.51&22.47&16.52&16.86&5.73&5.96 \\
	& SDA &25.89&25.70&23.81&24.15&21.18&21.55&17.84& 17.95\\
	& ReconNet &26.54&26.53&23.83&24.47&21.28&21.82&17.87&18.05\\
	& DR$^{2}$-Net &\textbf{29.42}& 29.63&\textbf{25.39}&\textbf{25.77}&\textbf{22.13}& \textbf{22.73}&\textbf{17.97}&\textbf{18.40} \\	

\hline
	& TVAL3~\cite{li2013efficient} & 27.77&27.77 &21.16 &21.16& 16.73& 16.73&11.09&11.11  \\
	& NLR-CS~\cite{dong2014compressive} & 25.91&26.06&14.59& 14.67&11.62 &11.97 &6.38&6.71  \\
Monarch	& D-AMP~\cite{metzler2014denoising} &26.39&26.55 &19.00 & 19.00&14.57 & 14.57&6.20&  6.20\\
	& SDA~\cite{mousavi2015deep} &23.54 & 23.32& 20.95& 21.04& 18.09& 18.19&15.31& 15.38 \\
	& ReconNet~\cite{kulkarni2016reconnet} & 24.31&25.06& 21.10& 21.51&18.19 & 18.32&\textbf{15.39}&15.49 \\
	& DR$^{2}$-Net &\textbf{27.95}& \textbf{28.31}&\textbf{23.10}&\textbf{23.56}&\textbf{18.93}&\textbf{19.23}&15.33&\textbf{15.50}\\	
\hline
	& TVAL3 & 27.17&27.24 &23.13&23.16&18.88&18.90&11.44&11.46  \\
	& NLR-CS & 26.53&26.72 &14.14 & 14.16& 10.59&10.92 &5.11& 5.44 \\
Parrot	& D-AMP &26.86 & 26.99&21.64 &21.64 & 15.78& 15.78&5.09& 5.09 \\
	& SDA & 24.48&24.36 &22.13&22.35 &20.37&20.67&17.70& 17.88\\
	& ReconNet &25.59 &26.22& 22.63&23.23&20.27 &21.06 &17.63&18.30 \\
	& DR$^{2}$-Net &\textbf{28.73} & \textbf{29.10}&\textbf{23.94}&\textbf{24.30}&\textbf{21.16}&\textbf{21.85}&\textbf{18.01}&\textbf{18.41} \\	
\hline
	& TVAL3 & 28.81&28.81 &23.86 &23.86&19.20&19.20&11.86&11.88  \\
	& NLR-CS & 29.11&29.27&14.82&14.86&10.76&11.21&5.38&5.72 \\
Boats	& D-AMP &29.26&29.26&21.95&21.95&16.01&16.01&5.34&5.34\\
	& SDA &26.56&26.25&24.03&24.18&21.29&21.54& 18.54&18.68\\
	& ReconNet &27.30&27.35&24.15&24.10&21.36&21.62&18.49&18.83 \\
	& DR$^{2}$-Net &\textbf{30.09}&\textbf{30.30}&\textbf{25.58}&\textbf{25.90}&\textbf{22.11}&\textbf{22.50}&\textbf{18.67}&\textbf{18.95}\\	
\hline
	& TVAL3 &\textbf{25.69}&25.70&21.91&21.92&18.30&18.33&11.97&12.00 \\
	& NLR-CS &24.88&24.96&14.18&14.22&11.04&11.43&5.98&6.31 \\
Cameraman	& D-AMP &24.41&24.54&20.35&20.35&15.11&15.11&5.64&5.64 \\
	& SDA &22.77&22.64&21.15&21.30&19.32&19.55&17.06& 17.19\\
	& ReconNet &23.15&23.59&21.28&21.66&19.26&19.72&\textbf{17.11}&\textbf{17.49}\\
	& DR$^{2}$-Net &25.62&\textbf{25.90}&\textbf{22.46}&\textbf{22.74}&\textbf{19.84}&\textbf{20.30}&17.08&17.34 \\
\hline
	& TVAL3 &35.42&35.54&28.69&28.74&20.63&20.65&10.97&11.01 \\
	& NLR-CS &\textbf{35.73}&\textbf{35.90}&13.54&13.56&9.06&9.44&3.91&4.25 \\
Foreman	& D-AMP &35.45&34.04&25.51&25.58&16.27&16.78&3.84&3.83 \\
	& SDA &28.39&28.89&26.43&27.16&23.62&24.09&20.07&20.23\\
	& ReconNet &29.47&30.78&27.09&28.59&23.72&24.60&20.04&20.33\\
	& DR$^{2}$-Net &33.53&34.28&\textbf{29.20}&\textbf{30.18}&\textbf{25.34}&\textbf{26.33}&\textbf{20.59}&\textbf{21.08} \\	
\hline
	& TVAL3 &32.08&32.13&26.29&26.32&20.94&20.96&11.86&11.90 \\
	& NLR-CS &\textbf{34.19}&\textbf{34.19}&14.77&14.80&10.66&11.09&4.96&5.29 \\
House	& D-AMP &33.64&32.68&24.84&24.71&16.91&17.37&5.00&5.02 \\
	& SDA &27.65&27.86&25.40&26.07&22.51&22.94&19.45&19.59\\
	& ReconNet &28.46&29.19&26.69&26.66&22.58&23.18&19.31&19.52\\
	& DR$^{2}$-Net &31.83&32.52&\textbf{27.53}&\textbf{28.40}&\textbf{23.92}&\textbf{24.70}&\textbf{19.61}& \textbf{19.99}\\	
\hline
	& TVAL3 &\textbf{29.62}&\textbf{29.65}&22.64&22.65&18.21&18.22&11.35&11.36 \\
	& NLR-CS &28.89&29.25&14.93&14.99&11.39&11.80&5.77&6.10 \\
Peppers	& D-AMP &29.84&28.58&21.39&21.37&16.13&16.46&5.79&5.85 \\
	& SDA &24.30&24.22&22.09&22.34&19.63&19.89&\textbf{16.93}&17.02\\
	& ReconNet &24.77&25.16&22.15&22.67&19.56&20.00&16.82&16.96\\
	& DR$^{2}$-Net &28.49&29.10&\textbf{23.73}&\textbf{24.28}&\textbf{20.32}&\textbf{20.78}&16.90& \textbf{17.11}\\			
\hline
			& TVAL3 & 27.84& 27.87&22.84&22.86&18.39 &18.40 &11.31&11.34  \\
		        &NLR-CS & 28.05 &28.19 &14.19 &14.22 &10.58 &10.98&5.30&  5.62\\
\textbf{Mean PSNR}	& D-AMP & 28.17& 27.67&21.14 &21.09 &15.49 &15.67 &5.19&5.23  \\
			& SDA &24.72& 24.55&22.43 &22.68 &19.96 &20.21 &17.29& 17.40 \\
			& ReconNet & 25.54& 25.92&22.68& 23.23&19.99 &20.44 &17.27&  17.55\\
			& DR$^{2}$-Net & \textbf{28.66}&\textbf{29.06} & \textbf{24.32}&\textbf{24.71}& \textbf{20.80}&\textbf{21.29}& \textbf{17.44}&\textbf{17.80}\\	
\hline
\end{tabular}
\label{Tab:compare_existing_methods}
\end{table*}
\section{Experiments} \label{sec:exp}
In this section, we conduct a series of experiments to test the reconstruction performance of Deep Residual Reconstruction Network (DR$^{2}$-Net).

\subsection{Implementation Details}
We use Caffe~\cite{jia2014caffe} to implement and train the DR$^{2}$-Net. The weights for $\mathcal{F}^{f}(\cdot)$ are initialised using gaussian distribution with standard variance 0.01. The weights of convolutional layers in DR$^{2}$-Net are initialised using gaussian distribution with standard variance 0.001. The batch-size is set as 128.

The experiments in Sect~\ref{Sect:LargeDataset} are conducted on the ImageNet Val dataset~\cite{imagenet_cvpr09}. The other experiments are conducted on the standard training and testing images described in~\cite{kulkarni2016reconnet}.

\subsection{Comparison with Existing Methods}

Firstly, we compare our DR$^{2}$-Net with five existing methods, \emph{i.e.,} TVAL3~\cite{li2013efficient}, NLR-CS~\cite{dong2014compressive}, D-AMP~\cite{metzler2014denoising}, SDA~\cite{mousavi2015deep}, and ReconNet~\cite{kulkarni2016reconnet}. The first three are iterative-based methods, and the last two are deep learning-based methods. The results are summarized in Table~\ref{Tab:compare_existing_methods}, where the best results are highlighted in bold.

As shown in Table~\ref{Tab:compare_existing_methods}, the proposed DR$^{2}$-Net obtains the highest mean PSNR values at four Measurement Rates (MRs). Compared with the existing methods, DR$^{2}$-Net has following advantages: 

1) Different from the existing deep learning-based methods, \emph{i.e.,} SDA~\cite{mousavi2015deep} and ReconNet~\cite{kulkarni2016reconnet}, DR$^{2}$-Net is more robust for the CS measurement at higher MRs. From Table~\ref{Tab:compare_existing_methods}, we could observe that the existing deep learning-based methods do not perform as good as DR$^{2}$-Net at higher MRs. For example at MR 0.25, the highest PSNR for deep learning-based methods is 25.92dB~\cite{kulkarni2016reconnet}, which is obviously lower than 28.19dB~\cite{dong2014compressive} of iterative-based method. Our DR$^{2}$-Net outperforms both the other deep learning-based methods and iterative-based methods. For instance, at MR 0.25, the DR$^{2}$-Net outperforms NLP-CS and ReconNet by 0.87dB and 3.14dB, respectively on PSNR.

2) Similar to the existing deep learning-based methods, DR$^{2}$-Net is robust for  the CS measurement at smaller MRs. As shown in Table~\ref{Tab:compare_existing_methods}, the iterative-based methods do not work well at smaller MRs, \emph{e.g.,} at MR 0.01 and 0.04. While the deep learning-based methods, \emph{e.g.,} SDA and ReconNet, both achieve higher mean PSNR values. However, DR$^{2}$-Net also constantly outperforms the other deep learning-based methods at lower MRs.
\begin{table*}
\small
\caption{Comparison of  loss and PSNR between DR$^{2}$-Net and ReconNet~\cite{kulkarni2016reconnet}. $fc_{1089}$ denotes the linear mapping network with one fully-connected layer containing $1089$ neurons. $Res_{n}$ denotes that further adding $n$ residual learning blocks to the linear mapping network. ``MR" denotes Measurement Rate.}
\centering
\begin{tabular}{|l|c|c|c|c|c|c|c|c|}
\hline
Models &\multicolumn{2}{|c|}{MR=0.01}&\multicolumn{2}{|c|}{MR=0.04}&\multicolumn{2}{|c|}{MR=0.10}&\multicolumn{2}{|c|}{MR=0.25}\\
\cline{2-9}
& Loss & PSNR/BM3D& Loss & PSNR/BM3D& Loss & PSNR/BM3D& Loss & PSNR/BM3D \\
\hline
\hline
ReconNet & 5.307 & 17.27/17.55& 2.34 & 19.98/20.44 & 1.139 & 22.6793/23.23 & 0.517 & 25.53/25.92 \\
\hline
$fc_{1089}$ & 5.42 &17.26/17.60  & 2.25 &20.05/20.46 & 1.01 & 23.10/23.39 & 0.398 &26.81/27.33 \\
\hline
$fc_{1089}$-$Res_{1}$ &5.2066&17.37/17.71 & 1.9946&20.59/21.03&0.7991&24.08/24.55&0.2880&28.41/28.81  \\
\hline
$fc_{1089}$-$Res_{2}$ & \textbf{5.1831}&17.40/17.79&1.9586&20.68/21.05&0.7756&24.28/24.71&0.2814& 28.62/29.01 \\
\hline
$fc_{1089}$-$Res_{3}$ &5.1896& \textbf{17.44}/\textbf{17.80}&1,9638&20.75/20.93& 0.7696&\textbf{24.32}/24.71&0.2801& 28.64/\textbf{29.09}  \\
\hline
$fc_{1089}$-$Res_{4}$ & 5.1983& 17.41/17.75&\textbf{1.9452}&\textbf{20.80}/\textbf{21.29}&\textbf{0.7650}& 24.25/\textbf{24.71}&\textbf{0.2788}& \textbf{28.66}/29.06\\
\hline
\end{tabular}
\label{Tab:recon_result}
\end{table*}
\subsection{Evaluation on Linear Mapping}

To verify that the linear mapping is able to reconstruct a high-quality preliminary image, we compare linear mapping with ReconNet~\cite{kulkarni2016reconnet} from two aspects: PSNR value and testing loss. Note that, the ReconNet contains one fully-connected layer and six convolutional layers, and our linear mapping network $\mathcal{F}^{f}(\cdot)$ only contains one fully-connected layer with 1089 neurons, denoted by $fc_{1089}$. The related results are summarized in Table~\ref{Tab:recon_result}.

From Table~\ref{Tab:recon_result}, we could see that the linear mapping network $fc_{1089}$ outperforms ReconNet both for PSNR and testing loss at MRs 0.25, 0.10, and 0.04, respectively. At MR 0.01, the testing loss for network $fc_{1089}$ is 5.42, which is larger than the 5.307 of ReconNet. However, the network $fc_{1089}$ achieves a comparable PSNR value with ReconNet, \emph{i.e.,} 17.26dB \emph{vs} 17.27dB of ReconNet. We further compare the time complexity between $fc_{1089}$ and ReconNet. As shown in Table~\ref{Tab:running_time}, the network $fc_{1089}$ is ten times faster than ReconNet. Therefore, we could conclude that the linear mapping could generate a reasonably good preliminary reconstruction image with fast speed.

 \begin{figure}
 \centering
\includegraphics[width=0.6\linewidth]{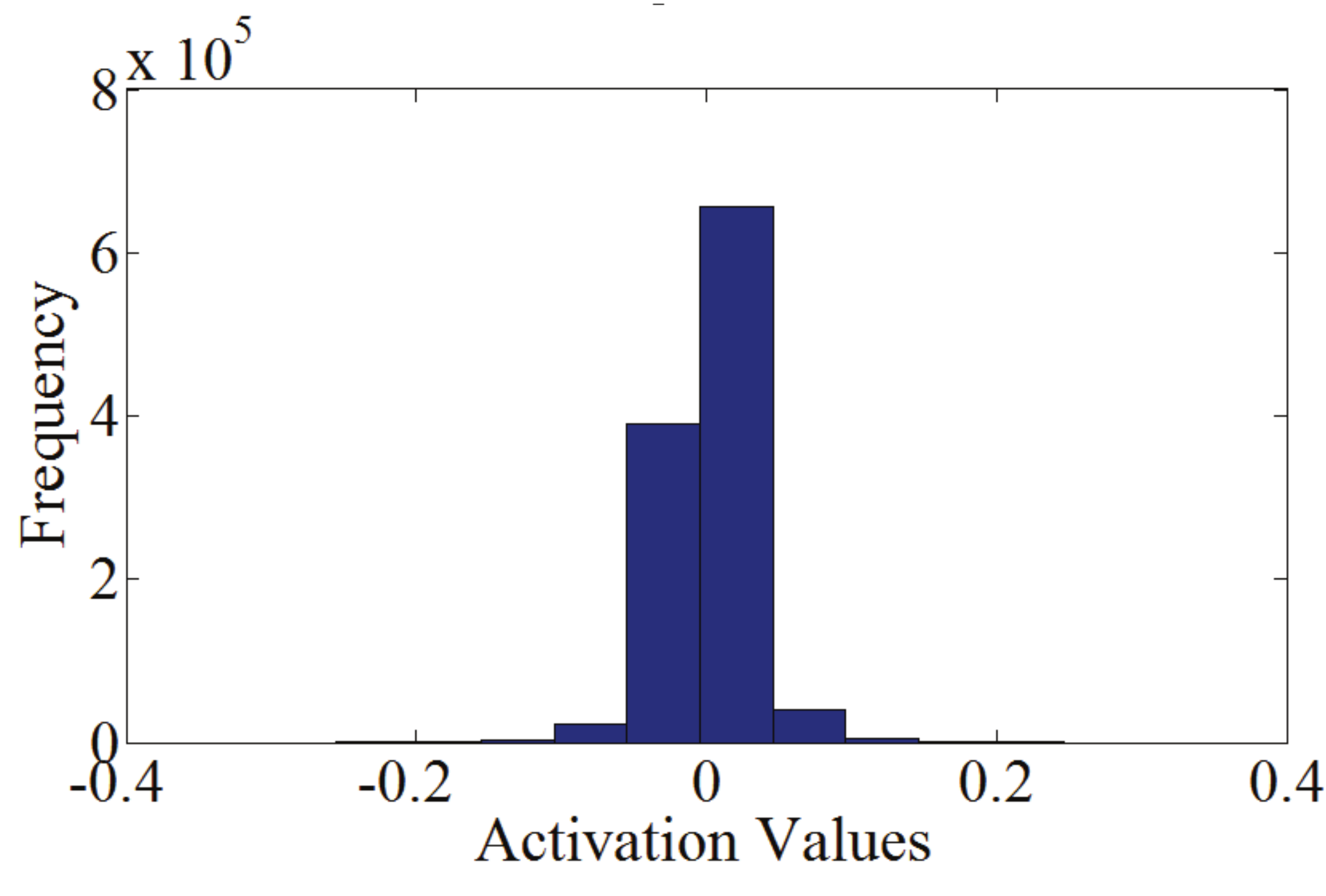}
\caption{Histogram of the activation values for residual learning.}
\label{Fig:hist}
\end{figure}

\subsection{Evaluation on Residual Learning}

As the linear mapping network has produced a preliminary reconstruction, the residual estimated by the residual learning should be small. As shown in Fig.~\ref{Fig:hist}, we could observe that residual learning network reasonably infers many small residual values, \emph{e.g.,} more than 95\% of the activation values are within the range of -0.05 to 0.05.

\begin{figure}
 \centering
\includegraphics[width=1\linewidth]{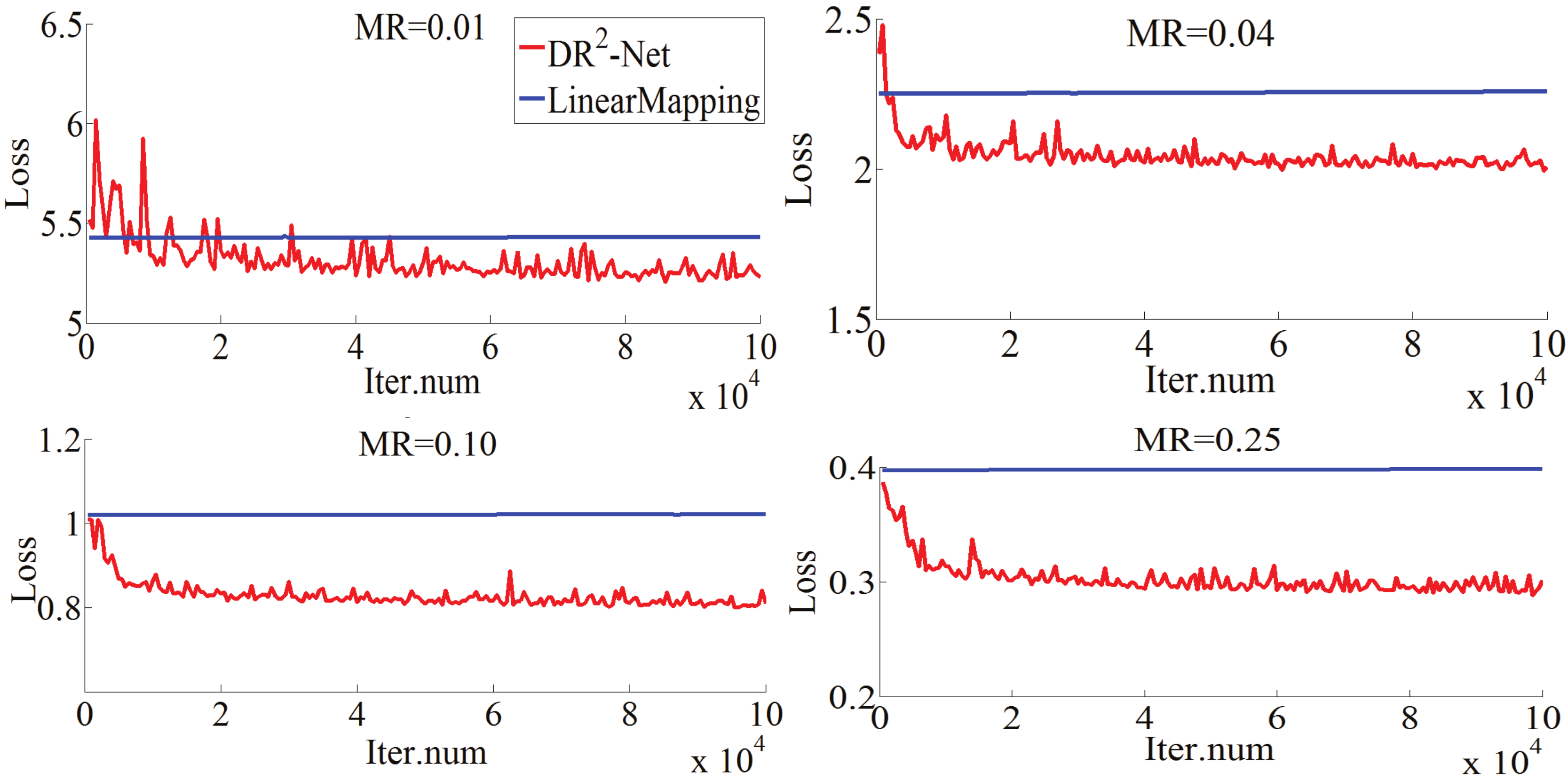}
\caption{The network losses on validation set during training procedure. The blue line and red line denote the loss of linear mapping network and DR$^{2}$-Net, respectively.}
\label{Fig:loss_variation}
\end{figure}

 The linear mapping network is first trained. Then, DR$^{2}$-Net is trained in an end-to-end manner. To show the validity of this training strategy, we analyze the network loss on the validation set during training, and summarize results in Fig.~\ref{Fig:loss_variation}. As shown in Fig.~\ref{Fig:loss_variation}, the loss for linear mapping network,\emph{ i.e.}, the blue solid lines, keeps stable during training. We also observe that DR$^{2}$-Net loss, \emph{i.e.}, the red lines, decreases as the iteration number increases. Note that, the DR$^{2}$-Net loss is composed of the linear mapping network loss and residual learning network loss. As the DR$^{2}$-Net loss keeps decreasing and linear mapping network loss is stable, it is easy to infer that the residual learning network is effectively optimized during the network training procedure.

Next, we show the effect of residual learning on reconstruction quality. As shown in Table~\ref{Tab:recon_result}, adding residual block substantially reduces the testing loss and improve the reconstruction quality. For example, compared to linear mapping network $fc_{1089}$, the deep residual network $fc_{1089}$-$Res_{1}$ achieves 0.09 dB, 0.54dB, 0.92dB, 1.6dB improvement for MR 0.01, 0.04, 0.10, 0.25, respectively.

We finally evaluate the performance of DR$^{2}$-Net with different number of residual learning blocks. As shown in Table~\ref{Tab:recon_result}, we could observe that introducing more residual learning blocks increases the PSNR performance and reduces the loss for most cases, \emph{e.g.,} using four residual blocks achieves the highest performance at MR 0.04, 0.10, 0.25.  It is also clear that the improvement becomes relatively smaller as more residual blocks are added to DR$^{2}$-Net. For example, when MR = 0.25, $fc_{1089}$-$Res_{4}$ only shows 0.25dB improvement over $fc_{1089}$-$Res_{1}$, which is smaller than 1.6dB improvement achieved by $fc_{1089}$-$Res_{1}$ over $fc_{1089}$. Therefore, residual learning is important for improving the reconstruction quality of DR$^{2}$-Net. However, it is not necessary to add too many residual blocks into the DR$^{2}$-Net. We add four residual blocks in DR$^{2}$-Net and use the network $fc_{1089}$-$Res_{4}$ as DR$^{2}$-Net.

\subsection{Time Complexity}

As the time complexity is a key factor for image compressive sensing, we analysis the time complexity of DR$^{2}$-Net. The related results are summarized in Table~\ref{Tab:running_time}. As the deep learning-based methods have been sped up 100 times over traditional iterative algorithms~\cite{kulkarni2016reconnet}, we only compare the time complexity between DR$^{2}$-Net and other deep-learning based methods. 

Firstly, we show the time complexity of DR$^{2}$-Net with different structures. From Table~\ref{Tab:running_time}, we could observe that the time complexity is linearity correlated with the depth of network, \emph{i.e.,} deeper DR$^{2}$-Net requires more running time. Among the five networks, the network $fc_{1089}$ contains only one fully-connected layer, thus shows the fastest speed, \emph{e.g.,} only needs $3$-$4$\emph{ms} to reconstruct a $256\times 256$ image. The deepest network $fc_{1089}$-$Res_{4}$ contains one fully-connected layer and 12 convolutional layers, and is about 30 times slower than $fc_{1089}$.

Next, we make comparison  with other deep learning-based methods: SDA\cite{mousavi2015deep} and ReconNet\cite{kulkarni2016reconnet}. Compared with SDA and ReconNet, the linear mapping network $fc_{1089}$ spends less time to achieve a comparable performance. Therefore, when the running time is an important concern, simply using network $fc_{1089}$ could be a good choice. From Table~\ref{Tab:running_time} and Table~\ref{Tab:recon_result}, we could observe that DR$^{2}$-Net with structure $fc_{1089}$-$Res_{1}$ not only achieves a faster speed, but also obtains a higher performance than ReconNet. As the other three DR$^{2}$-Nets contain more layers, they are slower than ReconNet.

\begin{table}
\small
\caption{Time (in seconds) for reconstructing a single $256\times 256$ image.}
\centering
\begin{tabular}{lcccc}
\hline
Models &MR=0.01&MR=0.04&MR=0.10&MR=0.25\\
\hline
\hline
SDA & 0.0045 & 0.0025 & 0.0029 & 0.0042 \\
ReconNet & 0.0326& 0.0301&0.0309 &0.0461 \\
\hline
\hline
$fc_{1089}$ &0.0028 &0.0032 &0.0037 &0.0041 \\
$fc_{1089}$-$Res_{1}$ &0.0190 & 0.0200&0.0198 & 0.0185\\
$fc_{1089}$-$Res_{2}$ &0.0344 &0.0315 & 0.0349& 0.0365\\
$fc_{1089}$-$Res_{3}$ &0.0508&0.0480 & 0.0433& 0.0505\\
$fc_{1089}$-$Res_{4}$ &0.0600 & 0.0576& 0.0565&0.0557 \\
\hline
\end{tabular}
\label{Tab:running_time}
\end{table}

 \subsection{Robustness to Noise}

\begin{figure}
 \centering
\includegraphics[width=1\linewidth]{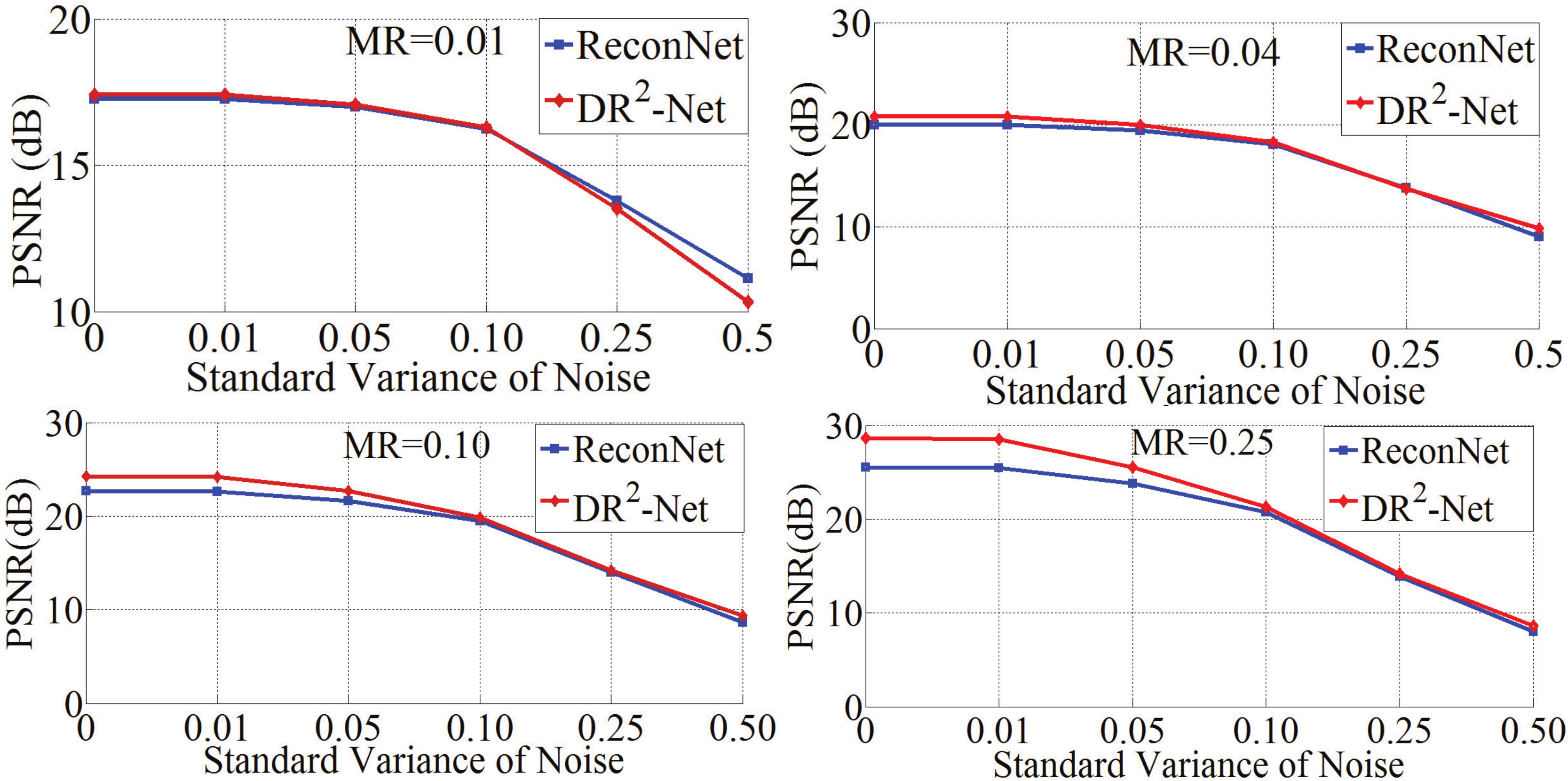}
\caption{Comparison of robustness to Gaussian noise between ReconNet and DR$^{2}$-Net.}
\label{Fig:noise}
\end{figure}

To show the robustness of DR$^{2}$-Net to noise, we investigate the reconstruction performance under the presence of measurement noise. We firstly add the standard Gaussian noise to the CS measurements of testing set,~\emph{i.e.,} we add five levels of noise corresponding to $\sigma$ = 0.01, 0.05, 0.1, 0.25 and 0.5, where $\sigma$  is the standard variance for the Gaussian noise. Then, the DR$^{2}$-Net trained on the noiseless CS measurements takes the noisy CS measurements as input, and outputs the reconstruction images. Aiming to make a fair comparison, we do not use BM3D~\cite{dabov2007image} to denoise the reconstruction images. The results are summarized in Fig.~\ref{Fig:noise}.

It could be observed that the DR$^{2}$-Net outperforms the ReconNet for the $\sigma$ = 0.01, 0.05, 0.1 at four MRs. For the cases that $\sigma$ = 0.25 and 0.5, DR$^{2}$-Net and ReconNet both do not work well because intense noises generated by large $\sigma$ significantly distract the original CS measurements.

\subsection{Scalability on Large-scale Dataset}~\label{Sect:LargeDataset}
 \begin{figure}
 \centering
\includegraphics[width=1\linewidth]{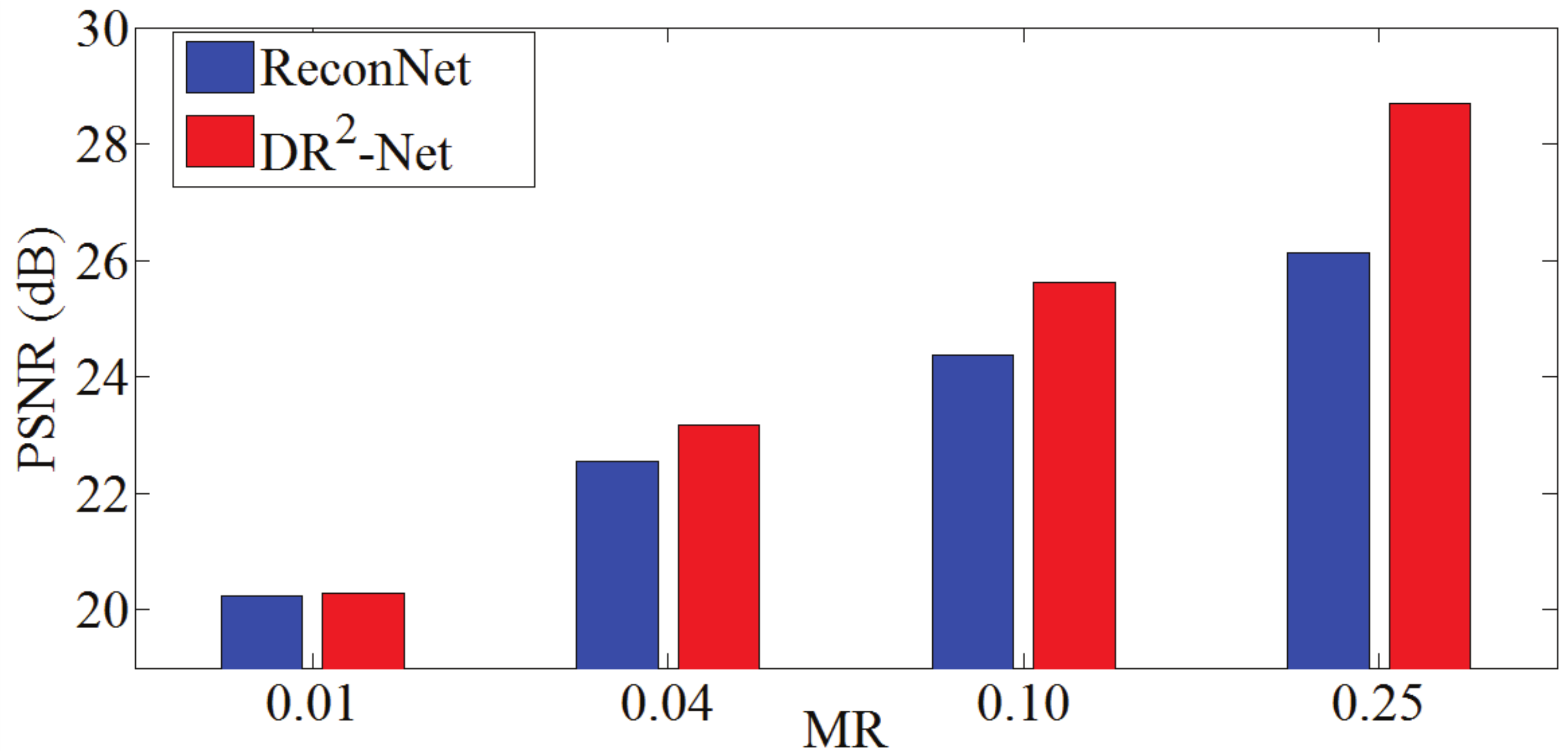}
\caption{Comparison between DR$^{2}$-Net and ReconNet on ImageNet Val dataset at four MRs.}
\label{Fig:imagenet_result}
\end{figure}

To verify the scalability of DR$^{2}$-Net, we conduct experiments to compare DR$^{2}$-Net with ReconNet on the large-scale ImageNet Val dataset~\cite{imagenet_cvpr09}, which contains 50,000 images from 1,000 classes. Note that, DR$^{2}$-Net and ReconNet are both trained based on the standard 91 images in~\cite{kulkarni2016reconnet}. 

As shown in Fig.~\ref{Fig:imagenet_result}, DR$^{2}$-Net gets better performance than ReconNet at four MRs. Especially for cases at higher MRs, DR$^{2}$-Net substantially outperforms ReconNet by large margins, \emph{e.g.,} DR$^{2}$-Net achieves nearly 3dB improvement over ReconNet at MR 0.25.

\begin{figure}
\small
\begin{center}
\subfigure[Images with high quality reconstruction.]{
\includegraphics[width=1\linewidth]{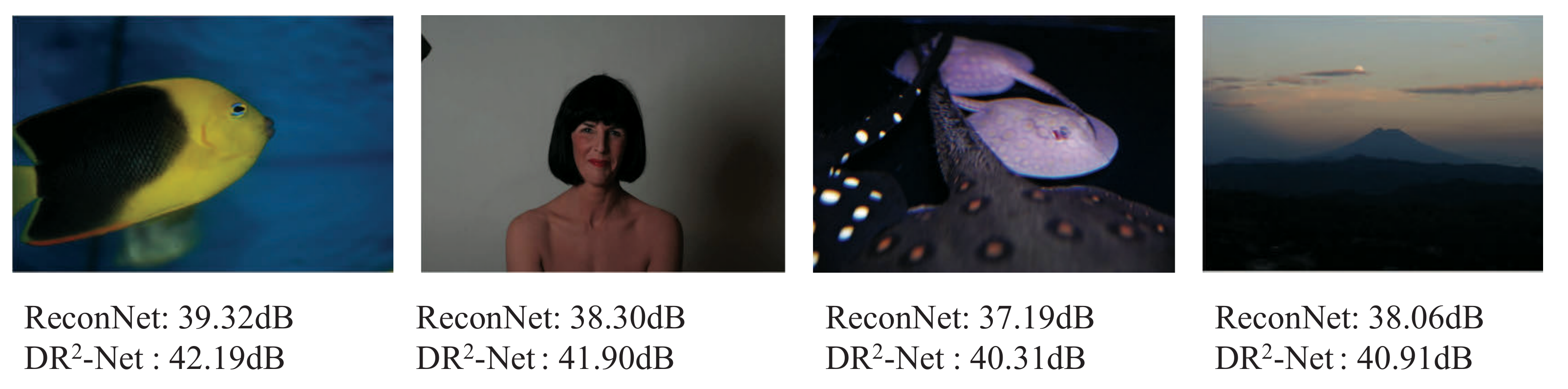}}
\subfigure[Images with middle quality reconstruction.]{
\includegraphics[width=1\linewidth]{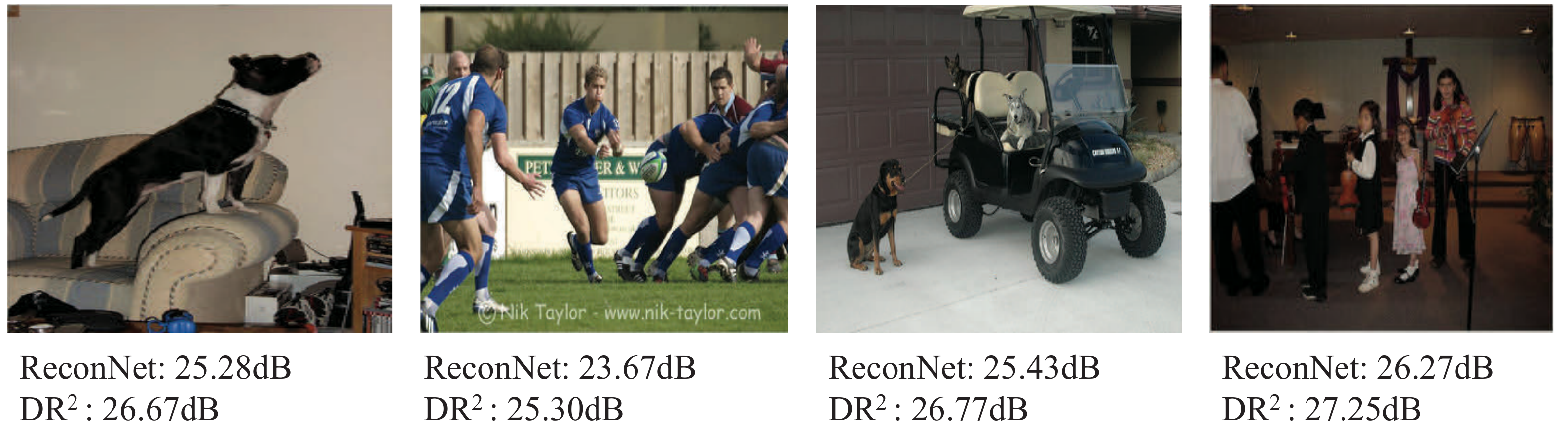}}
\subfigure[Images with low quality reconstruction.]{
\includegraphics[width=1\linewidth]{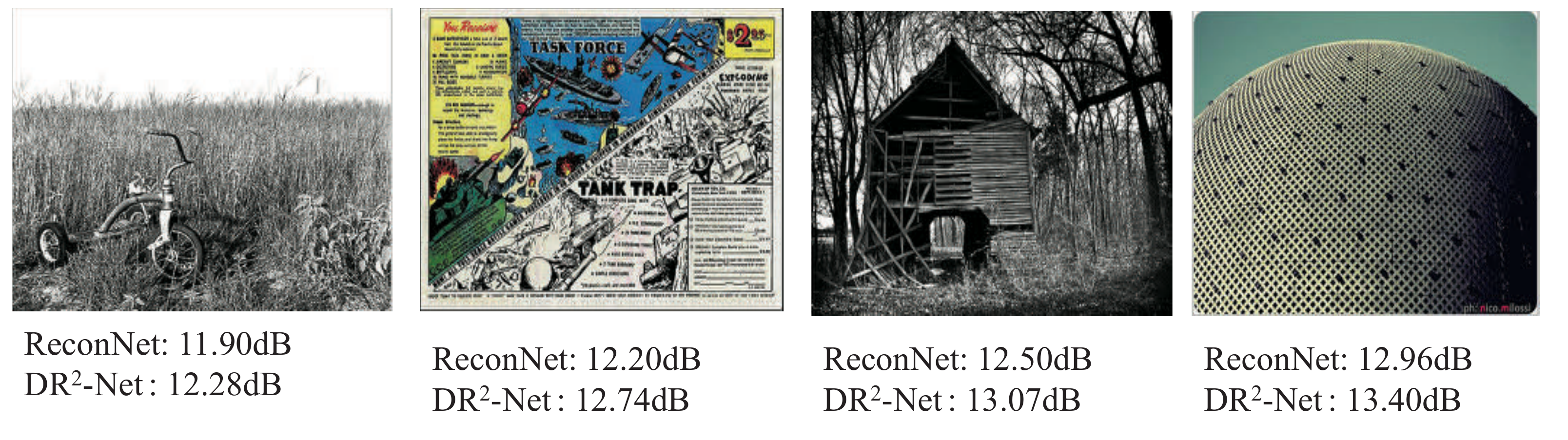}}
\end{center}
\caption{Images with different reconstruction quality. The results show the texture complexity affects the reconstruction quality of both DR$^{2}$-Net and ReconNet. The PSNR values are reported at MR 0.10.}
\label{Fig:imagenet}
\end{figure}

Fig.~\ref{Fig:imagenet} shows several testing images and the quality of their reconstruction by DR$^{2}$-Net and ReconNet, respectively at MR 0.10. It is clearly that DR$^{2}$-Net constantly outperforms ReconNet. We notice that images with higher reconstruction quality are generally smooth while those with lower reconstruction quality contain richer textures. Therefore, the texture complexity affects the construction quality of both DR$^{2}$-Net and ReconNet.

\begin{figure}[htb]
\small
\subfigure[Foreman]{
\includegraphics[width=0.9\linewidth]{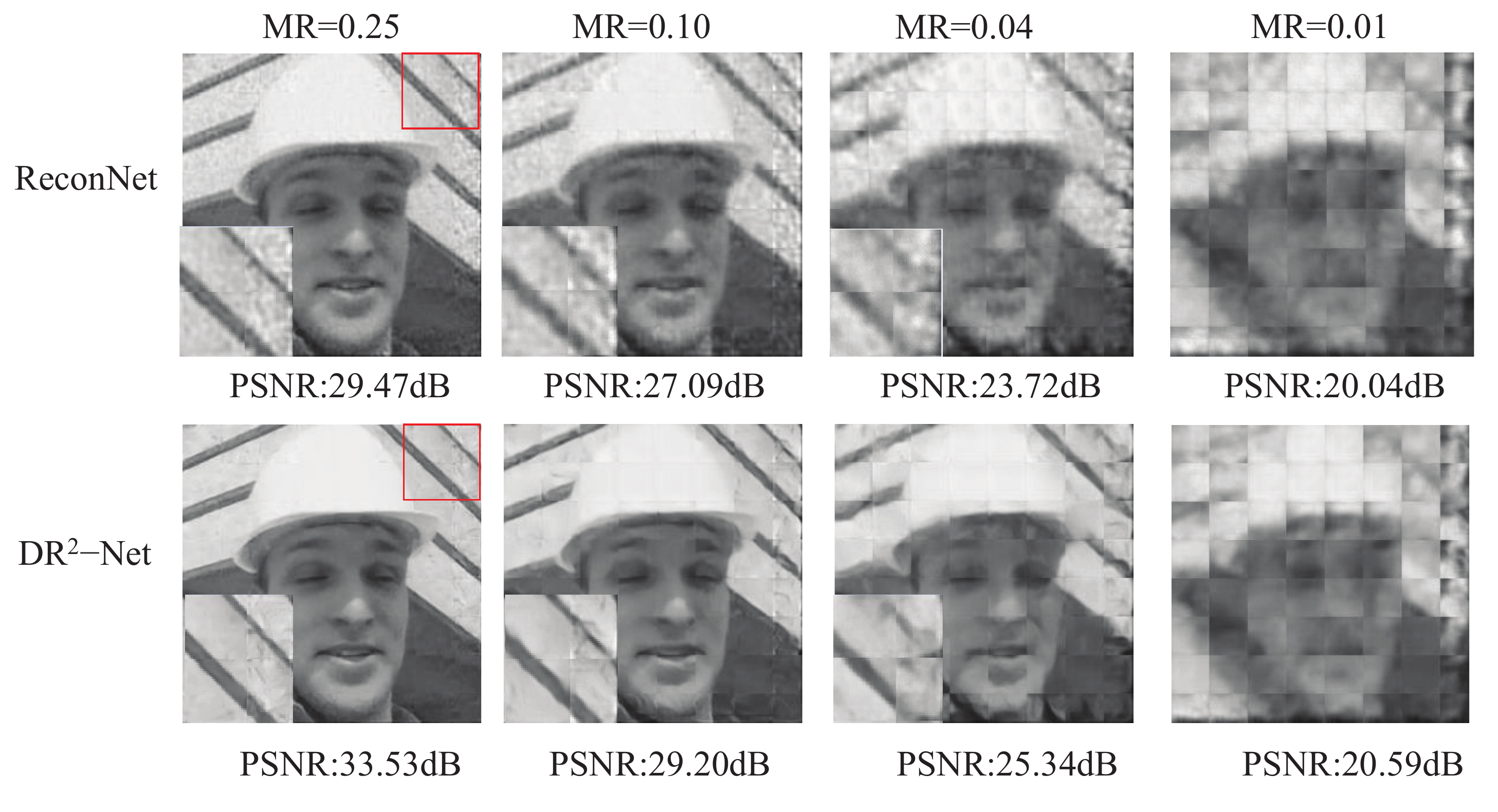}}
\subfigure[House]{
\includegraphics[width=0.9\linewidth]{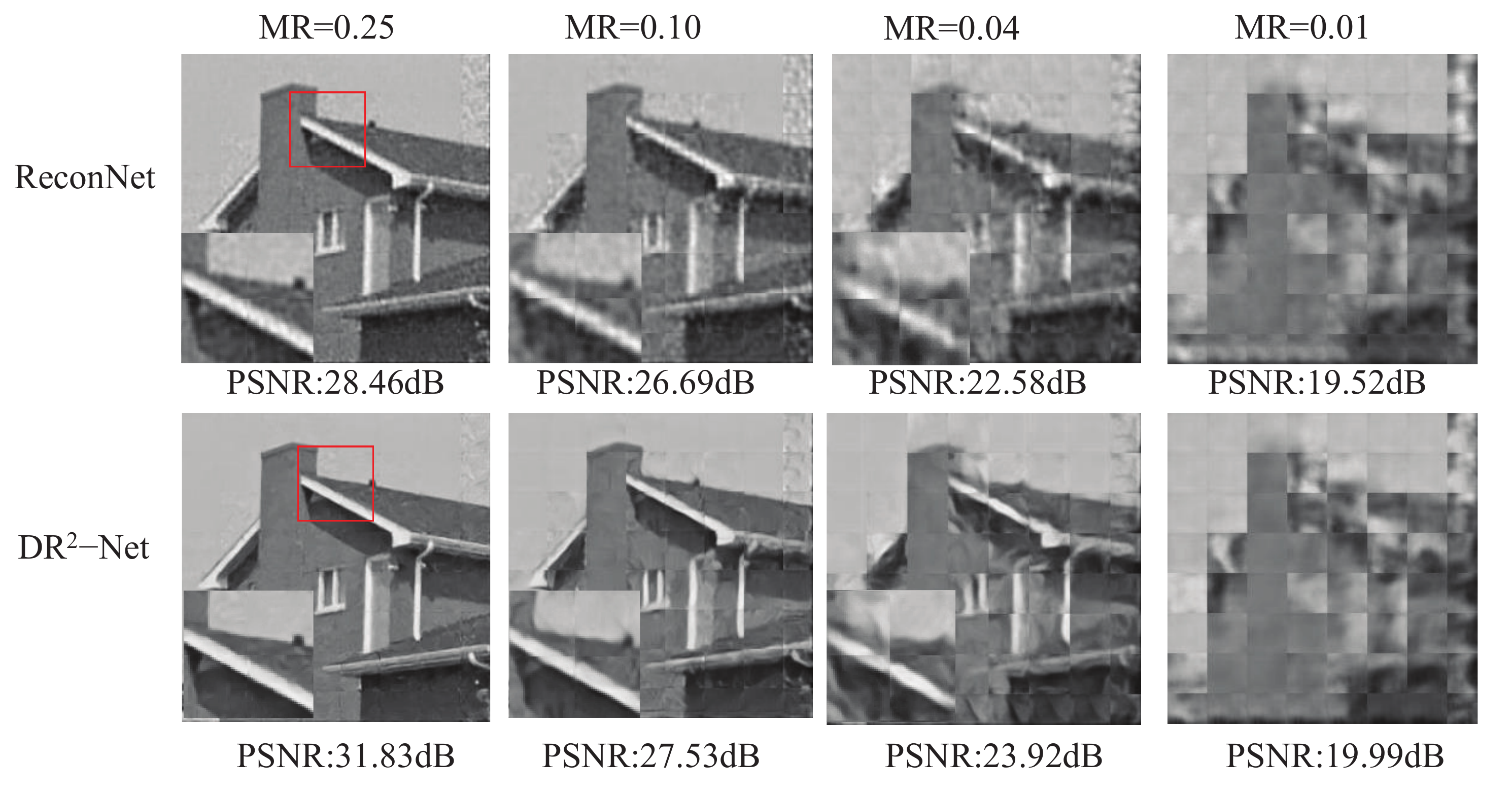}}\\
\subfigure[Peppers]{
\includegraphics[width=0.9\linewidth]{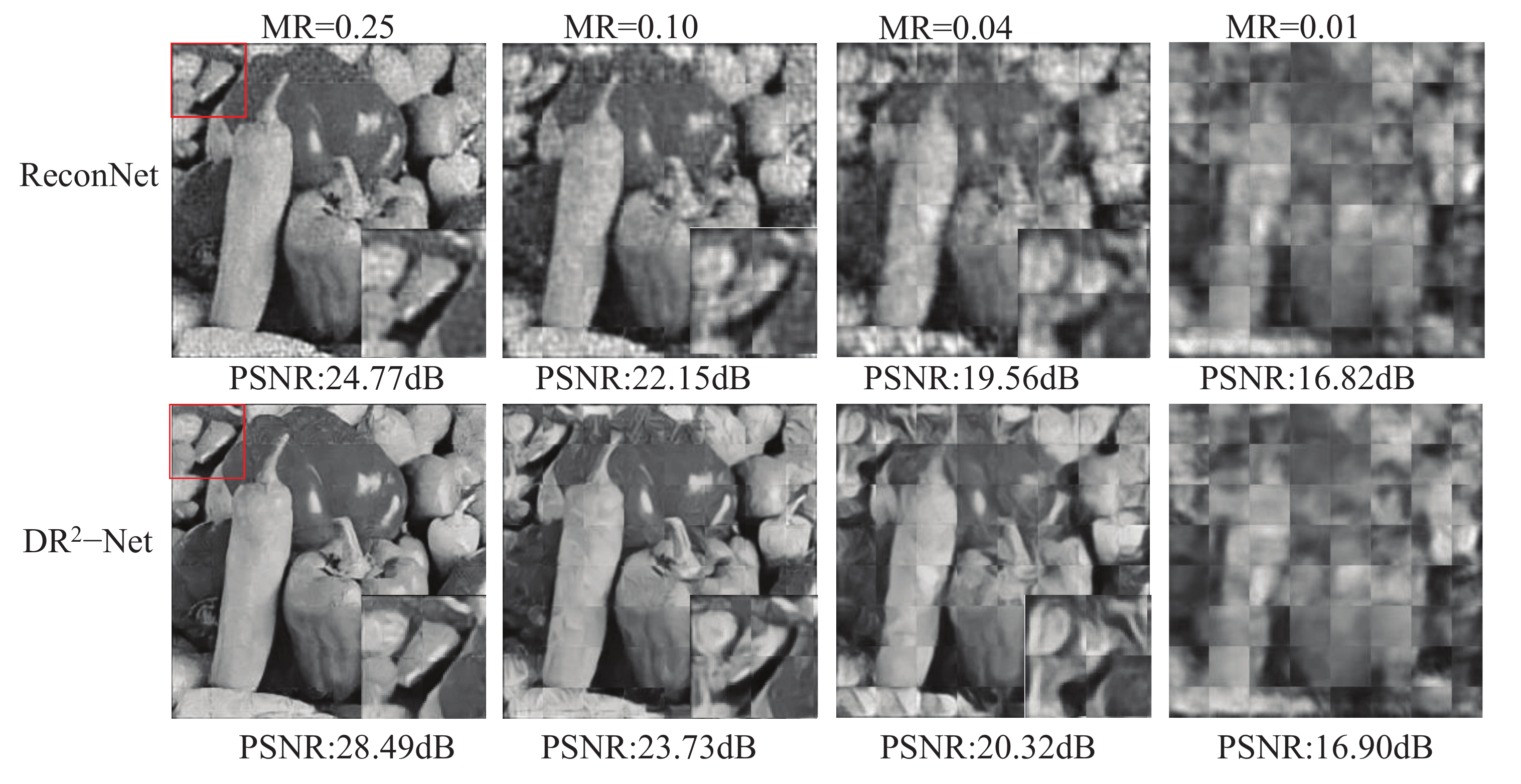}}
\caption{Reconstruction results from noiseless CS measurements without BM3D. It is obvious that DR$^{2}$-Net recovers more visually appealing images (Best viewed in color pdf).}
\label{Fig:reconstruction_results1}
\end{figure} 
 
\begin{figure*}
\small
\begin{center}
\subfigure[Barbara]{
\includegraphics[width=0.475\linewidth]{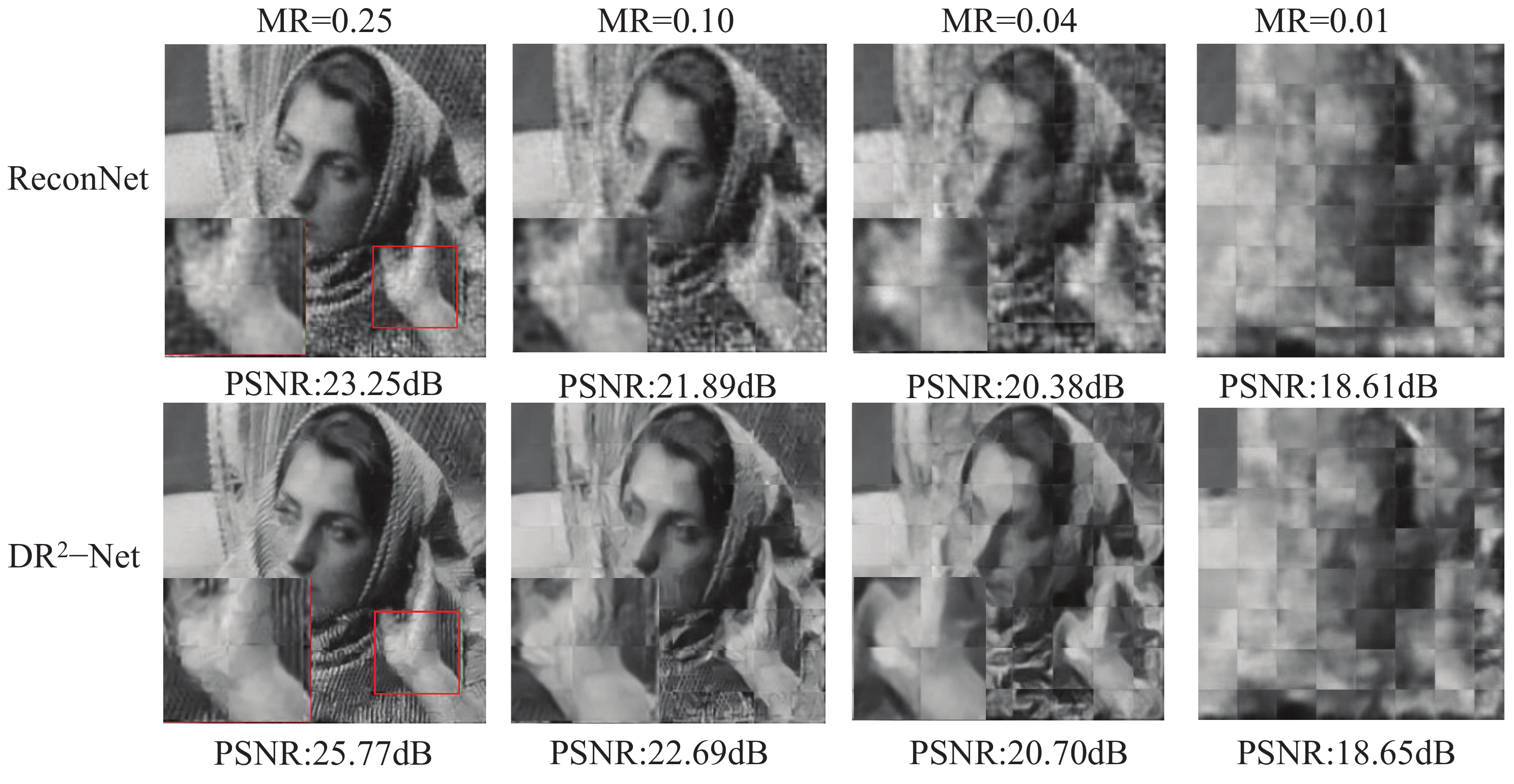}}
\subfigure[Fingerprint]{
\includegraphics[width=0.475\linewidth]{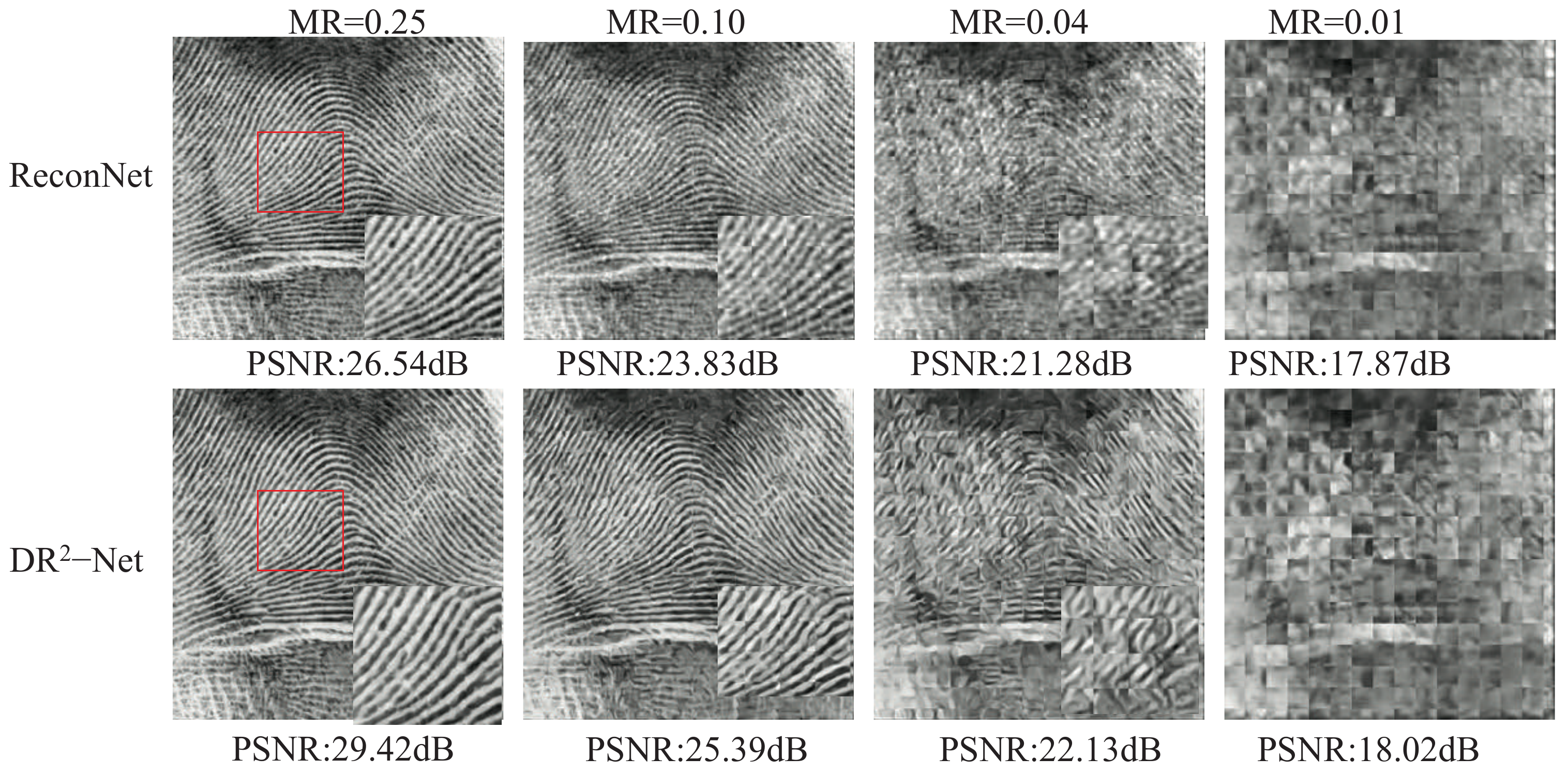}}\\
\subfigure[Flinstones]{
\includegraphics[width=0.475\linewidth]{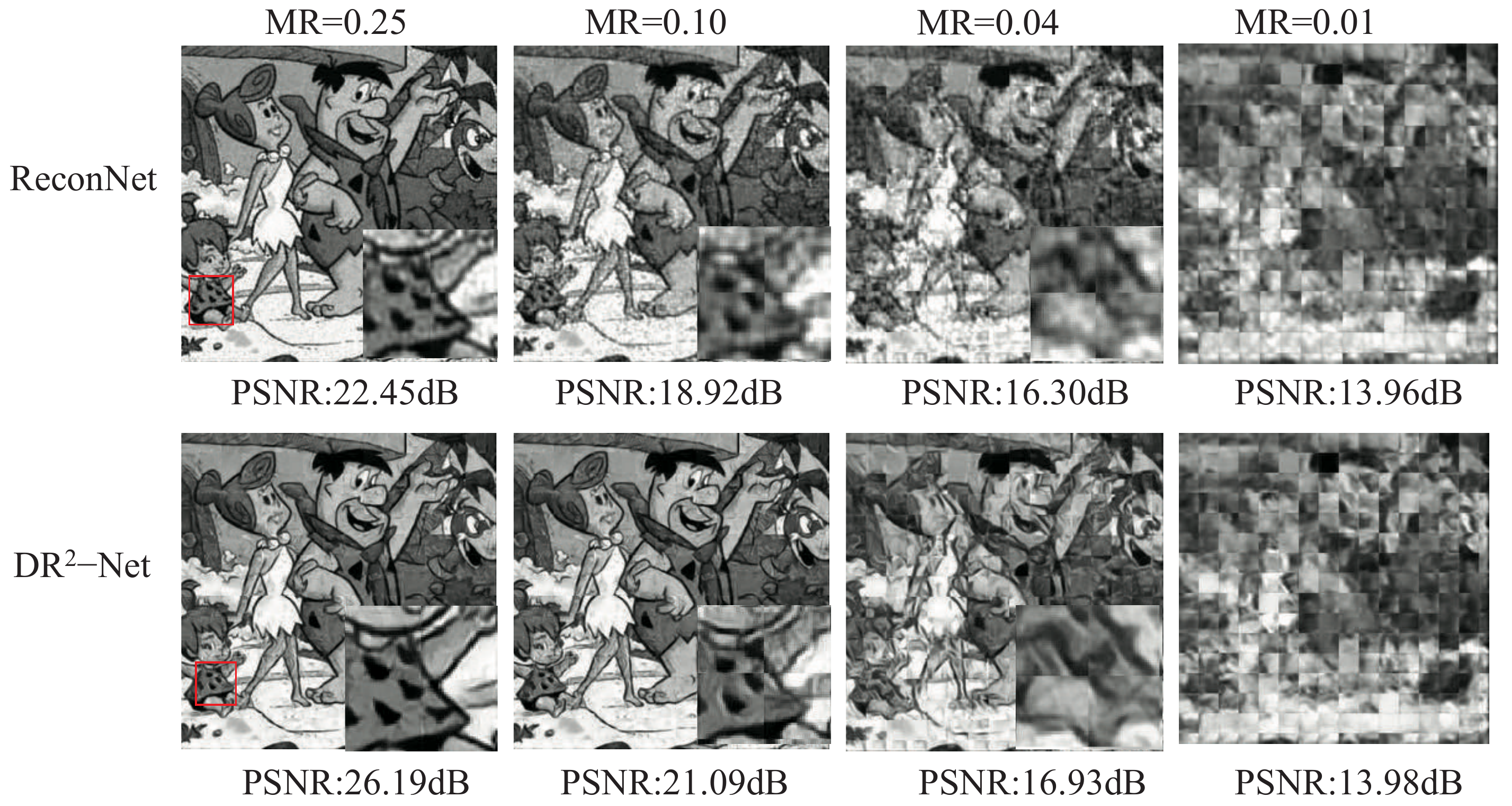}}
\subfigure[Lena]{
\includegraphics[width=0.475\linewidth]{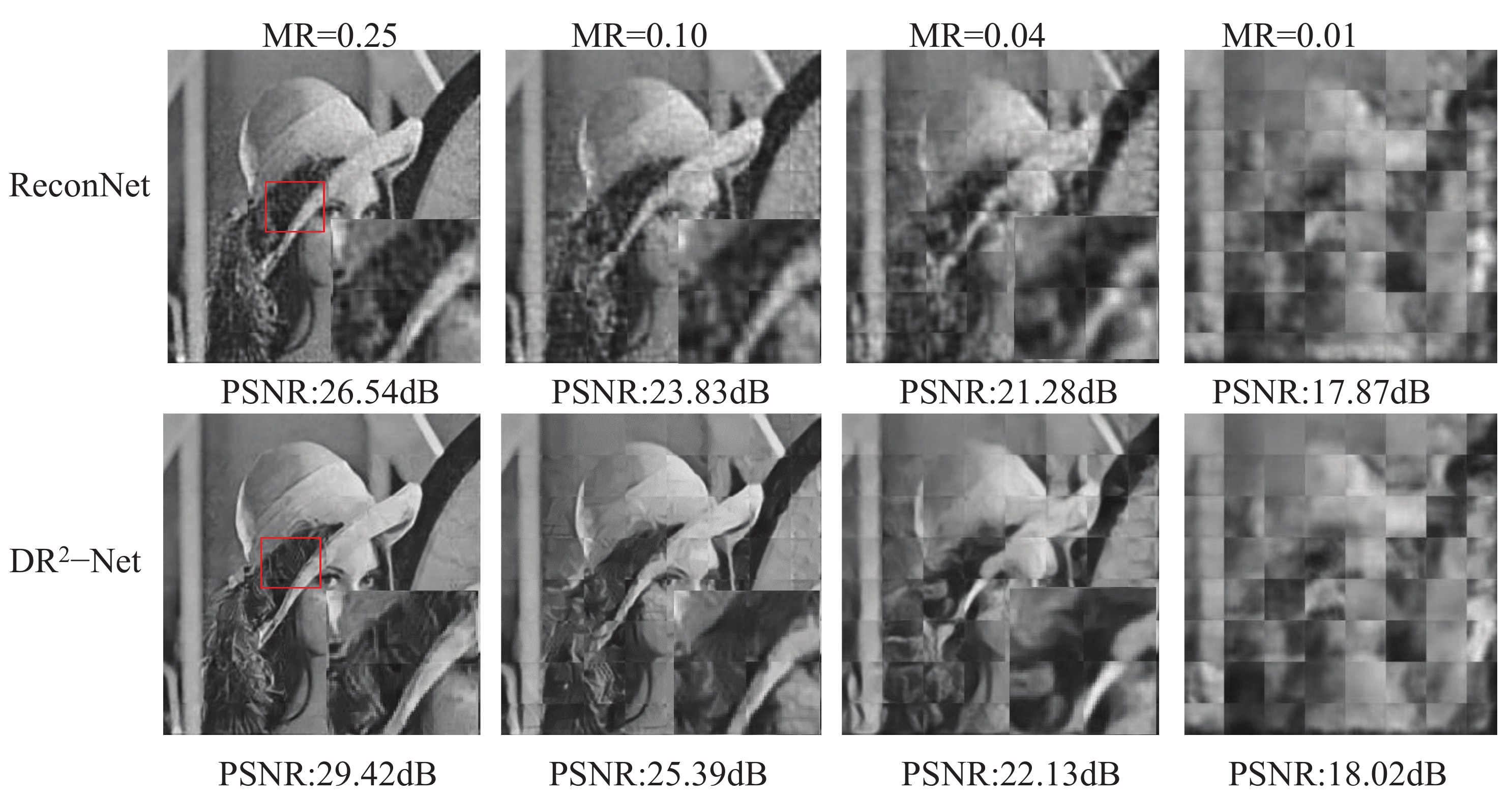}}\\
\subfigure[Monarch]{
\includegraphics[width=0.475\linewidth]{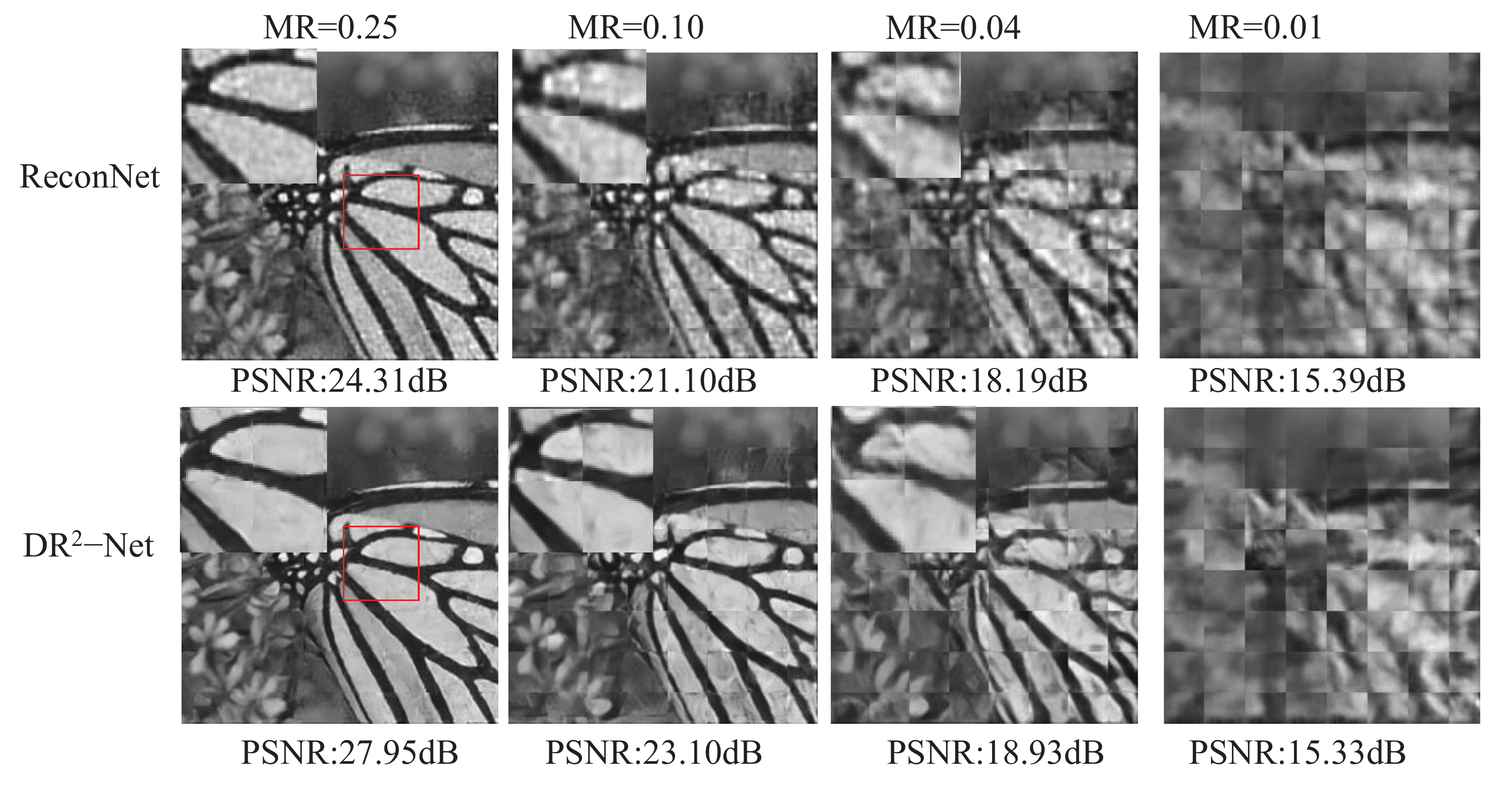}}
\subfigure[Parrot]{
\includegraphics[width=0.475\linewidth]{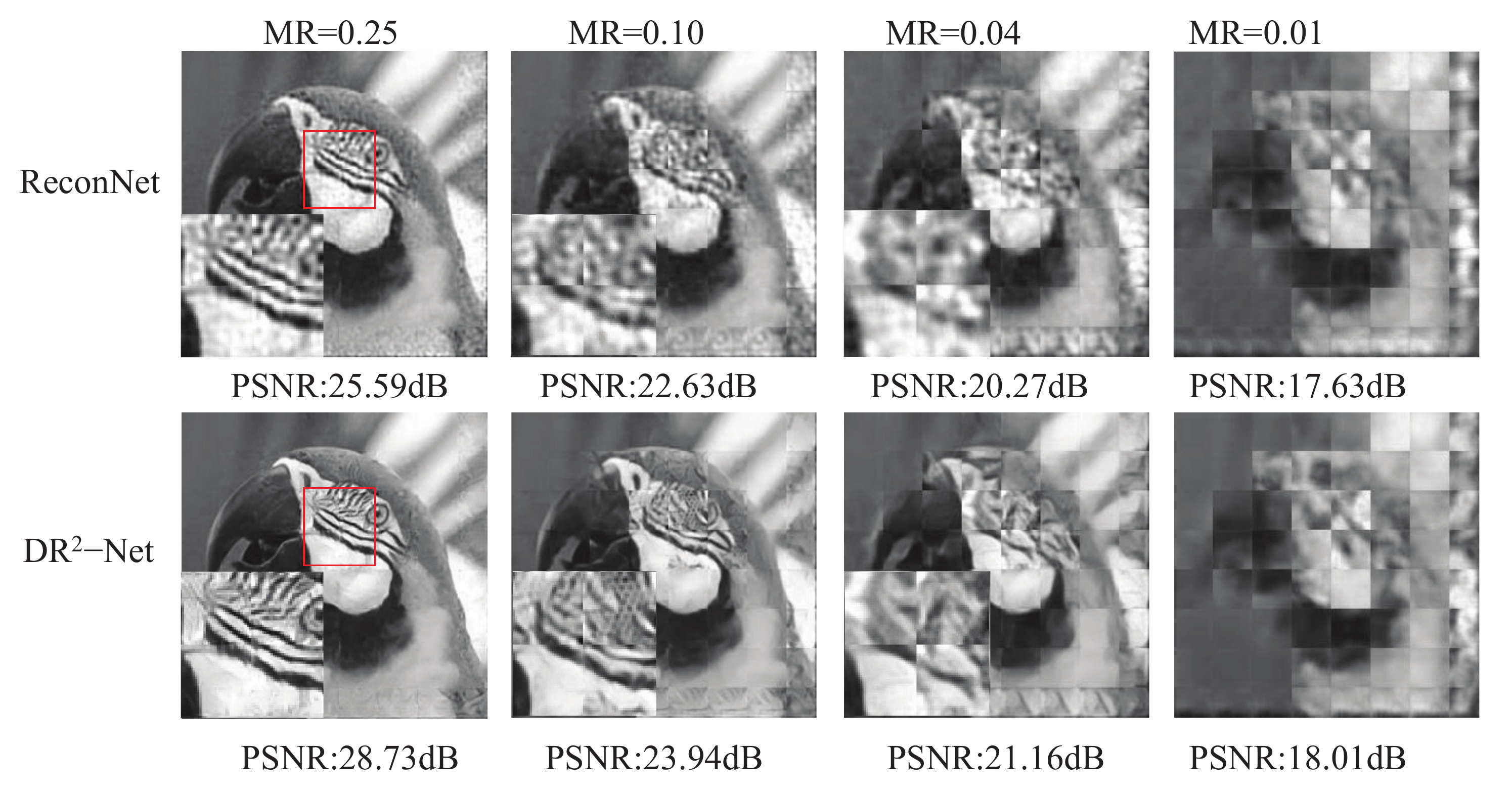}}\\
\subfigure[Boats]{
\includegraphics[width=0.475\linewidth]{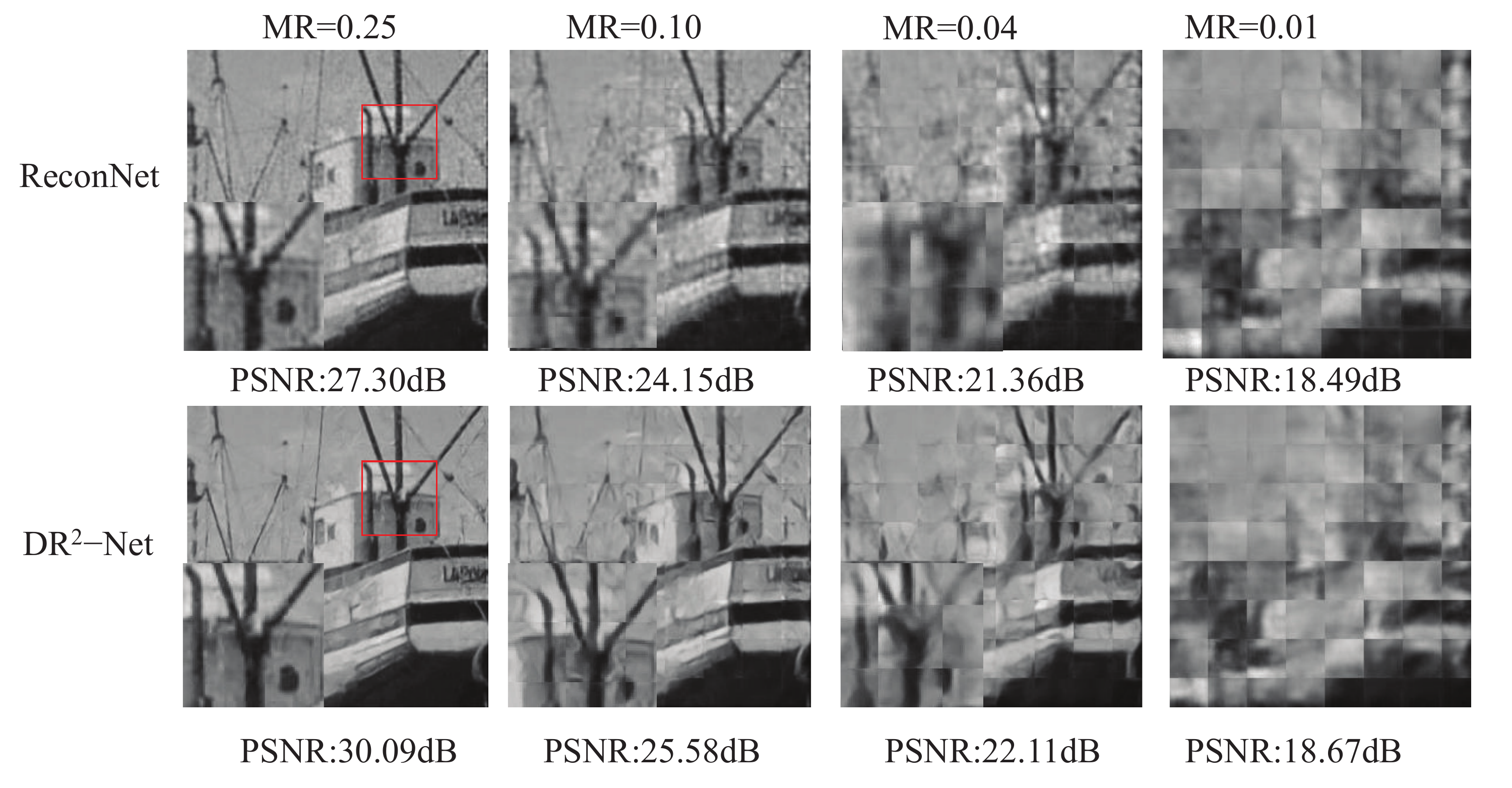}}
\subfigure[Cameraman]{
\includegraphics[width=0.475\linewidth]{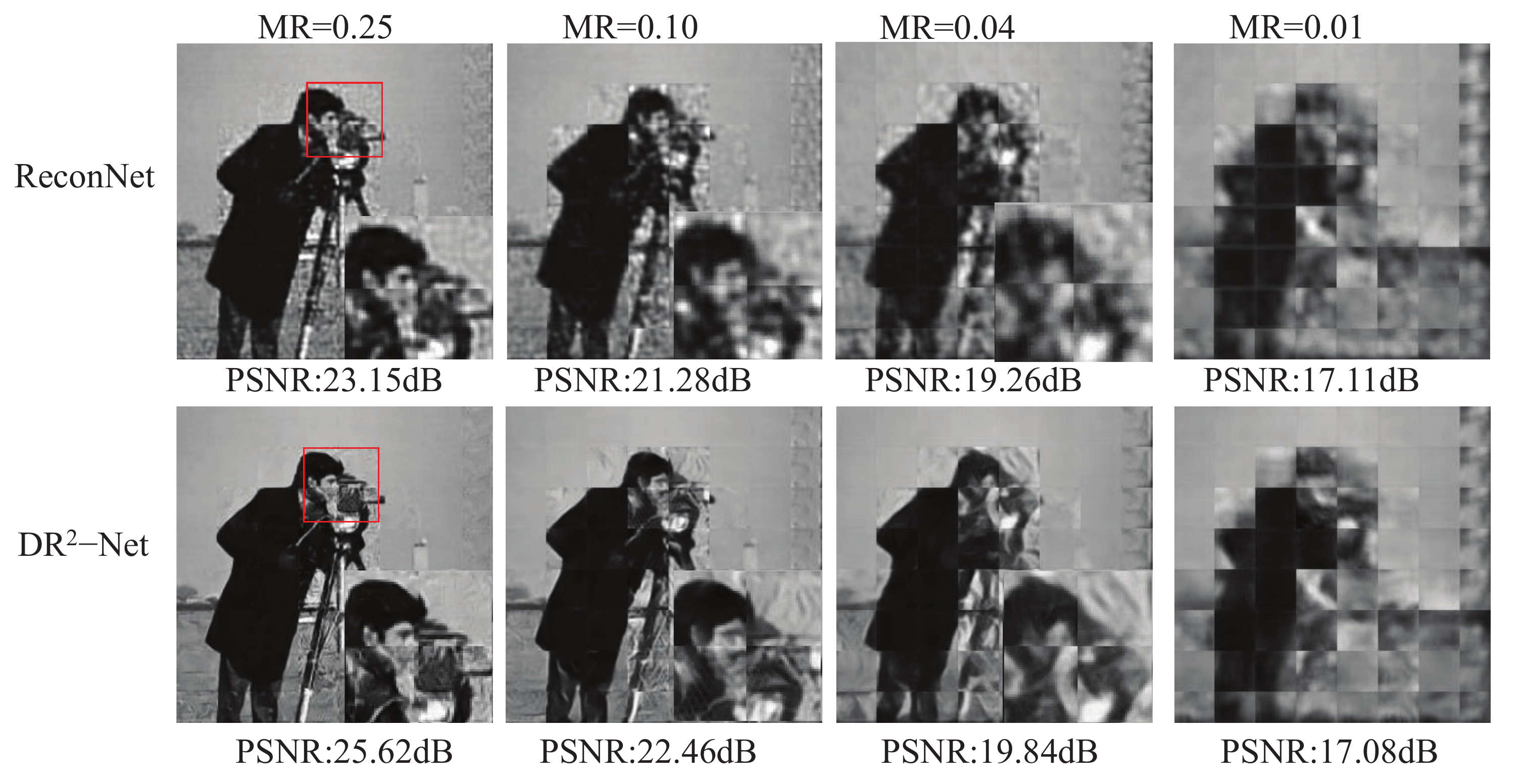}}\\
\end{center}
\caption{Reconstruction results from noiseless CS measurements without BM3D. It is obvious that DR$^{2}$-Net recovers more visually appealing images (Best viewed in color pdf).}
\label{Fig:reconstruction_results}
\end{figure*}

\section{Conclusion} \label{sec:conclusions}
Inspired by the existing works that deep learning-based methods could dramatically reduce the time complexity than iterative reconstruction algorithms. We propose a Deep Residual Reconstruction Network to reconstruct the image from its Compressively Sensed (CS) measurements. The proposed network is composed of a linear mapping network and several residual learning blocks. The fully-connected layer in neural network is taken as the linear mapping to produce the preliminary reconstruction. The residual learning blocks are trained to boost the preliminary result by inferring the residual between preliminary reconstruction and the groudtruth. Extensive experiments show that DR$^{2}$-Net outperforms the traditional iterative-based and deep learning-based methods by large margins in the aspects of both speed and quality. 


\ifCLASSOPTIONcaptionsoff
  \newpage
\fi

\bibliographystyle{IEEEtran}
\bibliography{IEEEabrv,egbib}

\end{document}